\documentclass{article}
\newcommand{\E}[1]{\mathbb{E}\left[ {#1} \right]}

\usepackage{microtype}
\usepackage{graphicx}
\usepackage{subcaption}
\usepackage{enumitem}

\usepackage{booktabs} % for professional tables
\usepackage{dsfont}
\usepackage{bm}
\usepackage{hyperref}
\usepackage{pgfplots}
\usepgfplotslibrary{fillbetween}
\pgfplotsset{compat = 1.3}
\definecolor{color0}{RGB}{108,91,224}
\definecolor{color1}{RGB}{197,53,120}
\definecolor{color2}{RGB}{233,105,44}
\definecolor{color3}{RGB}{234,172,62}
\definecolor{color4}{RGB}{000,000,000}

\usepackage[accepted]{icml2024}

\usepackage{amsmath}
\usepackage{amssymb}
\usepackage{mathtools}
\usepackage{amsthm}

\usepackage[capitalize,noabbrev]{cleveref}

\theoremstyle{plain}
\newtheorem{theorem}{Theorem}[section]
\newtheorem{proposition}[theorem]{Proposition}
\newtheorem{lemma}[theorem]{Lemma}

\theoremstyle{definition}
\newtheorem{definition}[theorem]{Definition}

\theoremstyle{remark}

\usepackage[textsize=tiny]{todonotes}

\icmltitlerunning{Reinforcement Learning and Regret Bounds for Admission Control}

\begin{document}

\twocolumn[
\icmltitle{Reinforcement Learning and Regret Bounds for Admission Control}

\icmlsetsymbol{equal}{*}

\begin{icmlauthorlist}
\icmlauthor{Lucas Weber}{lab}
\icmlauthor{Ana Bu\v{s}i\'c}{lab}
\icmlauthor{Jiamin Zhu}{comp}
\end{icmlauthorlist}

\icmlaffiliation{lab}{Inria and DI ENS, \'Ecole Normale Sup\'erieure, PSL University, Paris, France}
\icmlaffiliation{comp}{IFP Energies nouvelles, 1 et 4 avenue de Bois-Préau, 92852 Rueil-Malmaison, France}

\icmlcorrespondingauthor{Lucas Weber}{lucas.weber@inria.fr}
\icmlcorrespondingauthor{Ana Bu\v{s}i\'c}{ana.busic@inria.fr}
\icmlcorrespondingauthor{Jiamin Zhu}{jiamin.zhu@ifpen.fr}

\icmlkeywords{Reinforcement learning, regret minimization, admission control, ICML}

\vskip 0.3in
]

\printAffiliationsAndNotice{}  % leave blank if no need to mention equal contribution

\begin{abstract}
The expected regret of any reinforcement learning algorithm is lower bounded by $\Omega\left(\sqrt{DXAT}\right)$ for undiscounted returns, where $D$ is the diameter of the Markov decision process, $X$ the size of the state space, $A$ the size of the action space and $T$ the number of time steps. However, this lower bound is general. A smaller regret can be obtained by taking into account some specific knowledge of the problem structure. In this article, we consider an admission control problem to an $M/M/c/S$ queue with $m$ job classes and class-dependent rewards and holding costs. Queuing systems 
often have a diameter that is %at least 
exponential in the buffer size $S$, making the previous lower bound prohibitive for any practical use. We propose an algorithm inspired by UCRL2, and use the structure of the problem to upper bound the expected total regret by $O(S\log T + \sqrt{mT \log T})$ in the finite server case. In the infinite server case, we prove that the dependence of the regret on $S$ disappears.
\end{abstract}

\section{Introduction}
In the admission control problem, jobs
arrive at the entrance of a finite buffer queue, and the system decides whether to accept or reject a job based on the filling status of the queue and the job's class. The arrivals of jobs follow  independent Poisson processes with class-specific rates. Accepted jobs generate class-dependent rewards, while rejected jobs leave without waiting. The system also incurs class-dependent holding costs when accepted jobs that are not served immediately wait for service.
% a server to be freed.

Our goal is to minimize the expected total regret, which measures the difference between the maximum reward that could have been achieved and the actual collected rewards. 
For a concrete application, minimizing the regret or maximizing the average reward
asymptotically yield an optimal policy. 
However, by minimizing the regret, 
we control the quality of transient policies.
Minimizing the regret allows to better manage jobs appearing during the learning 
and limits the loss of revenue for the system operator due to the exploration.

The admission control problem is highly relevant in communication systems that experience losses due to limited buffer space or service capacity.
To ensure uninterrupted data flow for critical traffic types, like voice or streaming media, while still providing best-effort service for noncritical services, such as web traffic or file transfers, these jobs are categorized into priority classes in Quality of Service (QoS)-differentiated services \citep{carpenterDifferentiatedServicesInternet2002,zhouReinforcementLearningbasedAdaptive2017}. Admission control protocols seek to find a balance by accommodating as many jobs as possible while ensuring preferential treatment for higher-priority jobs.
Priority can be modeled through adequate definitions of rewards and holding costs.
Similar balancing issues can be found in various fields, 
one example being electric vehicles charging with different priorities \cite{alsabbaghDistributedChargingManagement2019}. 

With the increasing success of reinforcement learning for industrial applications, it has now become important to study not only the asymptotic optimality, but also regret minimization for admission control. 

\subsection{Related work}

The admission control literature is rich with articles interested in finding structural properties of the policies maximizing the long-term average reward.
Feinberg and Yang \citeyearpar
{feinbergOptimalityTrunkReservation2011}
extended Haviv and Puterman's work \citeyearpar{havivBiasOptimalityControlled1998}
and Lewis et al.'s article \citeyearpar{lewisBIASOPTIMALITYQUEUE1999} by considering several job classes
and class-dependent holding costs.
This corresponds to our setting, 
except that we minimize the regret 
and that we learn the arrival rates.
Feinberg and Yang have characterized optimal, bias optimal, and Blackwell optimal policies, and have adapted Miller's Policy Iteration (PI) \citeyearpar{millerQueueingRewardSystem1969a} which yields an optimal average reward policy
to the continuous-time admission control problem.

These works have inspired algorithms that use the structural properties of an optimal policy.
 For instance, Massaro et al. \citeyearpar{massaroOptimalTrunkReservationPolicy2019} proposed the Integer Gradient Ascent algorithm that leverages the knowledge of the threshold structure to accelerate the search for an optimal policy. They based their proposition on an earlier work by Feinberg and Reiman \citeyearpar{feinbergOptimalityRandomizedTrunk1994a} proving that, without holding costs, there exists an optimal threshold policy where the thresholds define the number of jobs in the queue above which jobs are rejected.
Following the same trend, Roy et al. \citeyearpar{royOnlineReinforcementLearning2022}
proposed another structure-aware two-timescale continuous time algorithm, similar to an Actor-Critic algorithm, accepting a holding cost that depends on the total number of jobs in the system.
In this paper, we leverage the special structure of PI for admission control, highlighted by Feinberg and Yang \citeyearpar{feinbergOptimalityTrunkReservation2011}.

Regret minimization in reinforcement learning (RL)
aims to maximize the cumulative reward gained by an agent during its learning process.
For general Markov decision processes (MDP),
the expected total regret
of any learning algorithm
is lower bounded by
$\Omega(\sqrt{DXAT})$\footnote{$f(n) \in \Omega(h(n))$ if there exists a positive constant $C$ and rank $n_0$ such that for all $n\geq n_0$, $0 \leq C h(n) \leq f(n)$},
where $D$ is the diameter of the MDP,
$X$ is the size of the state space,
$A$ is the size of the action space
and $T$ is the number of steps 
taken \citep{JMLR:v11:jaksch10a}.
Note that for a queue of length $S$, $X=S+1$ and $A=2^m$ where $m$ is the number of job classes.

Jaksch et al's Upper Confidence Reinforcement Learning (UCRL2)
algorithm has been designed to handle the exploration-exploitation trade-off efficiently.
It is an improvement of Auer and Ortner's UCRL \citeyearpar{auerLogarithmicOnlineRegret2006}, 
which is itself an extension to RL of the Upper Confidence Bound algorithm (see \citealp[Chapter 7]{lattimoreBanditAlgorithms2020} for a detailed presentation) that was developed for multi-armed bandit problems.
UCRL2 is a model-based algorithm
that estimates confidence bounds for the reward of each state-action pair
and for the probability transition matrix of each action. 
These estimations allow the derivation of a confidence set for the underlying MDP. The agent selects actions that optimize the return of the optimistic MDP within that confidence set, encouraging exploration of uncertain or underexplored state-action pairs. As the agent receives feedback, the confidence bounds are updated to accurately reflect the learned information. UCRL2's regret is upper bounded by $\tilde{O}(DX\sqrt{AT})$
\footnote{
$f(n)\in \tilde{O}(h(n))$ is equivalent to: $\exists k, f(n)\in O\left(h(n) \log^k(h(n))\right)$.
}
with high probability, and other algorithms have later reached an upper bound of $\tilde{O}(\sqrt{DXAT})$ 
\citep{zhangRegretMinimizationReinforcement2019a, tossouNearoptimalOptimisticReinforcement2019}.

Structured decision processes,
%such as admission control problems 
in queuing systems
may come with tighter bounds, as shown by
Anselmi et al. \citeyearpar{anselmiReinforcementLearningBirth2022}.
They have proposed an algorithm
and an upper bound on the expected regret
for the specific case of an $M/M/1/S$ queue
with controlled service rate
and an energy minimization objective.
Their upper bound is asymptotically independent of the number of states
and the diameter of the MDP.

In the context of admission control, Cohen et al. \cite{cohenLearningBasedOptimalAdmission2024} have studied the asymptotic regret of a $M/M/1$ queue with a single class of job.

In this paper we consider a multi-class admission control problem and propose a finite-time bound on the expected regret. We consider an $M/M/c/S$ queue with $m$ job classes and class-dependent rewards and holding costs. 
The operator does not know the arrival rates.
We assume that the number of servers and the service distribution are known. 
In most admission control applications, this assumption holds since the servers are owned by the system operator.
Moreover, we assume that the operator knows the rewards and holding costs of each job class.

\subsection{Contributions}

\begin{enumerate}[label=(\roman*), wide, labelwidth=!, labelindent=0pt]
    \item
%Our contributions are as follows.
%(i) 
% {\color{red}We show that Feinberg and Yang's \citeyearpar{feinbergOptimalityTrunkReservation2011}  results for admission control with state-independent arrival rates hold when the arrival rates depend on the state of the system.}
We extend Feinberg and Yang's \citeyearpar{feinbergOptimalityTrunkReservation2011} results to state-dependent arrival rates.
This allows to extend UCRL2's optimism principle \cite{JMLR:v11:jaksch10a} to queues with class-dependent holding costs, for which the priority order of jobs can change with the number of jobs in the system.
% {\color{red}: the optimistic CTMDP can have state-dependent arrival rates.}
\item 
%(ii) 
% {\color{red}We introduce the UCRL for admission control algorithm (UCRL-AC) with holding costs that depend on the job classes
% and limits the expected regret by $O(S\log T + \sqrt{mT \log T})$ when the number of servers is finite and the immediate rewards $R^{(i)}, i=1, \ldots, m$ are bounded.
% A significant advantage of our algorithm lies in its use of the structure of the problem. 
% It allows us to employ a PI algorithm, which produces a linear dependence on $S$ by exploiting bias results that are specific to gain optimal policies.
% Moreover, in the infinite-server case,
% this dependence on $S$ completely vanishes.}
We introduce the UCRL for admission control algorithm (UCRL-AC) with class-dependent holding costs. We show that its expected regret is upper bounded by $O(S\log T + \sqrt{mT \log T})$ when the number of servers is finite and the immediate rewards $R^{(i)}, i=1, \ldots, m$ are bounded.
In the infinite-server case,
this dependence on $S$ completely vanishes.
By using PI, our algorithm can exploit the structure of optimal policies to obtain a linear dependence on $S$ through results on the bias that are specific to gain optimal policies.
Note that in classical UCRL algorithms, the upper bounds on regret depend on the diameter $D$ of the MDP, and $D$ is exponential in $S-c$ in our admission control problem
(see appendix \ref{appendix: diameter lower bound}) for a finite number of servers $c<S$.
\item 
%(iii) 
By using the structure of the problem to reformulate Policy Iteration, 
our algorithm reduces its memory use to $O(mS)$, while previous implementations would require a memory space of $O(S^2 A)$. 
\item 
%(iv) 
We obtain closed-form expressions of the relative bias for a given policy $\pi$ that we leverage to 
reduce the time complexity of PI.
\item 
%(v) 
UCRL-AC can also work with Value Iteration (VI), which is known to converge geometrically fast, at the cost of an
exponential dependence on the queue size $S$ for the regret bound at worst. 
\end{enumerate}

The article is structured as follows. Section \ref{section: problem formulation} presents 
the admission control problem.  Section \ref{section: background} provides essential background information. Section \ref{section: algorithm} describes UCRL-AC. In section \ref{section: computing the bias}, we show how to accelerate the computation of the bias function. We present the upper bound of the expected regret in section \ref{section: regret}. 
Experimental evaluations are conducted in section \ref{section: experiments}, and we conclude in section \ref{section: conclusion}. 
Detailed proofs of our results can be found in the appendix.

\section{Problem Formulation}
\label{section: problem formulation}
\subsection{Admission Control Problem}
We consider the admission control problem in an $M/M/c/S$ queue
with $c$ identical servers, finite capacity $S$, and $m$ job classes %that operates 
under the first-in-first-out rule.
Jobs of each class $i=1,\dots, m$ arrive at the system
according to independent Poisson processes with arrival rate $\lambda^{(i)}>0$.
We note $\Lambda = \sum_{i=1}^m \lambda^{(i)}$ the global arrival rate,
and we suppose $\Lambda \in [\Lambda_{\min}, \Lambda_{\max}]$. 
Thus, an arrival has a probability $p^{(i)}=\lambda^{(i)}/\Lambda$ of being of class $i$.
Upon arrival, the job class is revealed.
Each server has a service rate $\mu \in ]0, +\infty[$, 
so that the time to serve a job follows an exponential distribution of the parameter $\mu$.
When $s$ jobs are in the system,
the total service rate is $\mu(s)=\min(s,c)\mu$.

When the queue is full, % with $S$ jobs, 
new arrivals are rejected.
Otherwise, a controller decides whether to accept or reject the arriving job.
If a server is free when a job enters the queue, it is served immediately.
Otherwise,
the job waits in the queue.

Upon admitting a class $i$ job,
the system collects an immediate non-negative reward $R^{(i)}$
and incurs a non-negative holding cost $C(W(s))$ which depends on
the random waiting time $W(s)$ spent in the system until joining a free server. 
The expected reward for admitting job class $i$ when $s$ jobs are already in the system is
\begin{equation*}
\label{eq: definition r_i(s)}
    r^{(i)}(s)=R^{(i)} -\E{C(W(s))} 
\end{equation*}
For instance, if the holding cost $C(t)$ is proportional to the waiting time $t$, i.e. $C(t)=\gamma t$, then $r^{(i)}(s)=R^{(i)}-\gamma \E{W(s)}$ where $W(s)$ for $s\geq c$ is an Erlang random variable with parameters $(s-c+1, c\mu)$. 
 
We suppose that the service rate $\mu$ is known.
The arrival rates $\lambda^{(i)}$, $1 \leq i \leq m$ are unknown.

\subsection{Birth-and-Death Process}
We formulate our problem as a CTMDP.
When there are $s$ jobs in the queue,
we say that the system is in state $s$.
Let us denote $\mathcal{X}=\{0, 1, \dots, S\}$ the state space
and $\mathcal{I}=\{1, 2, \dots, m\}$ the set of job classes.
An action $a$ represents the set of job classes to be admitted in the current state:
$a\in \mathcal{A}=\mathcal{P}(\mathcal{I})$ where $\mathcal{P}(\mathcal{I})$ is the power set of $\mathcal{I}$.
When choosing action $a$ for state $s$,
the effective birth rate in state $s$ becomes $\Lambda(s,a)= \sum_{i \in a} \lambda^{(i)}$
and the expected reward per unit of time is $R(s,a) = \sum_{i\in a} \lambda^{(i)} r^{(i)}(s)$. 
We recall that $\mu(s)=\min(s,c)\mu$ and we note $\mu_{\max}= \mu(S)$.

Since $\mathcal{X}$ and $\mathcal{A}$ are finite, there exists
a deterministic stationary optimal policy. % exists.
In this article, we focus on deterministic stationary policies and denote by $\Pi$ the family of all such policies. 
For a policy $\pi\in \Pi$ that selects action $\pi(s)$ in state $s\in \mathcal{X}$,
the CTMDP becomes a continuous-time Markov chain that can be represented as a birth-and-death process. In state $s$, with policy $\pi$, the birth rate $\Lambda^\pi(s)$ and the expected reward per unit of time $R^\pi(s)$ are given below:
\begin{equation}
    \Lambda^\pi(s) = \sum_{i \in \pi(s)} \lambda^{(i)} \qquad
    R^\pi(s) = \sum_{i\in \pi(s)} \lambda^{(i)} r^{(i)}(s)
\end{equation}

The infinitesimal generator for the CTMDP under policy $\pi$ is the matrix $Z^\pi$ which is determined by $\Lambda^\pi(s) $
 and $\mu(s)$ as follows: $(Z^\pi)_{s, s+1}=\mathds{1}_{s<S}\Lambda^\pi(s)$, $(Z^\pi)_{s, s-1}=\mathds{1}_{s>0}\mu(s)$, $(Z^\pi)_{s,s}=-\mathds{1}_{s>0}\mu(s)-\mathds{1}_{s<S}\Lambda^\pi(s)$ and $(Z^\pi)_{i,j}$ is equal to zero everywhere else. 
 
 We recall that the long-run average reward of a policy $\pi$ is defined as
\begin{equation*}
    \rho^\pi = \underset{T\to\infty}{\lim}\frac{1}{T} \E{\int_0^TR^\pi(s_t) dt} 
\end{equation*}
where $s_t$ is the state of the system at time $t$. 

Our objective is to minimize the expectation of the regret $\Delta(T)$ defined as
\begin{equation*}
    \Delta(T) = T\rho^* - \sum_{s,i} \nu_T(s, i) r^{(i)}(s)
\end{equation*}
where $\rho^*$ is the long-run average reward per unit of time for a gain optimal policy (see definition \ref{def: optimal policy}),
and $\nu_T(s,i)$ is the number of jobs of class $i$ that are accepted in state $s$ until time $T$.

\section{Background}
\label{section: background}

\subsection{Bellman Equation}
For a policy $\pi$, let us note the infinitesimal generator $Z^\pi$ and the vector of rewards per second $R^\pi$ in each state. The average reward per second $\rho^\pi$ and the bias $h^\pi$ satisfy the following matrix equation 
\citep[equation (4)]{millerQueueingRewardSystem1969a}
\begin{equation}
    \label{eq: fixed-point equation}
    \rho^\pi = R^\pi + Z^\pi h^\pi
\end{equation}
For any policy $\pi$, there is only one recurrent class, so that the CTMDP is unichain and for all state $s$, $\rho^\pi(s)=\rho^\pi$. We call $\nabla h^\pi(s) = h^\pi(s) - h^\pi(s+1)$ the relative bias.
As stated by Feinberg and Yang \citeyearpar{feinbergOptimalityTrunkReservation2011}, for the admission control problem, the matrix equation \eqref{eq: fixed-point equation} becomes: for all $s$,
\begin{equation}
    \label{eq: average reward for admission control}
    \rho^\pi = \sum_{i\in \pi(s)} \lambda^{(i)} \left(r^{(i)}(s) - \nabla h^\pi(s)\right) + \mu(s) \nabla h^\pi(s-1)
\end{equation}
This equation is well defined since for $s=0$, $\mu(s)=0$ and for $s=S$, no job is accepted.

We recall the definition of gain optimal policies and the Bellman equation satisfied by a gain optimal policy.
\begin{definition}[Gain optimal policy]
    \label{def: optimal policy}
    A policy $\pi_g$ is gain optimal if its average reward $\rho^{\pi_g}$ is optimal:
\begin{equation*}
    \forall \pi \in \Pi, \rho^{\pi} \leq \rho^{\pi_g}=\rho^*
\end{equation*}
\end{definition}

A policy $\pi$ satisfying the following Bellman equations is gain optimal: for all state $s$,
\begin{equation}
\label{eq: Bellman equation}
    \rho^{\pi} = \underset{a \in \mathcal{P}(\mathcal{I})}{\max} \sum_{i \in a} \lambda^{(i)} \left( r^{(i)}(s) - \nabla h^{\pi}(s) \right) + \mu(s) \nabla h^{\pi}(s-1)
\end{equation}

\subsection{Policy and Value Iteration}

Feinberg and Yang have proposed an adaptation of Miller's PI for the admission control problem with state- and class-dependent reward functions. The evaluation step consists in solving equation \eqref{eq: average reward for admission control}, while the improvement step selects a new policy $\pi'$ such that for all states $s$, 
\begin{equation}
    \label{eq: policy improvement}
    \pi'(s)=\underset{a\in \mathcal{P}(\mathcal{I})}{\arg \max} \sum_{i\in a} \lambda^{(i)} (r^{(i)}(s)-\nabla h^\pi(s))
\end{equation} 
Evaluation and improvement alternate until the policy cannot be improved anymore. PI yields at convergence a policy $\pi$ that is a fixed point of equation \eqref{eq: policy improvement}. Thus, $\pi$ satisfies the Bellman equation \eqref{eq: Bellman equation} and is therefore gain optimal. The corresponding algorithm is presented in Alg. \eqref{alg: PI}.

\begin{center}
\begin{algorithm}
    \caption{PI for Admission Control}
    \begin{algorithmic}
        \STATE \textbf{Evaluation}: compute $\rho^{\pi_l}$ and $h^{\pi_l}$ by solving equation \eqref{eq: average reward for admission control} for $s\in\{0, 1, \ldots, S\}$.

        \STATE \textbf{Improvement}: compute new policy $\pi_{l+1}$ satisfying for all $s=0, \ldots, S-1$:
        \begin{equation*}            
        \forall\, 1\leq i \leq m, \begin{cases}
                r^{(i)}(s)>\nabla h^{\pi_l}(s) &\implies i \in \pi_{l+1}(s)\\
                r^{(i)}(s)<\nabla h^{\pi_l}(s) &\implies i \notin \pi_{l+1}(s)
            \end{cases}
        \end{equation*}
    \end{algorithmic}
    \label{alg: PI}
\end{algorithm}
\end{center}

The admission control problem naturally defines a CTMDP. 
Since VI is designed for MDPs, it must be applied to the equivalent uniformized formulation.
We make repeated use of the general CTMDP uniformization method presented by Puterman in section 11.5.1 of his book \citeyearpar{putermanMarkovDecisionProcesses2005a}. 
Let us note $P_U(a)$ the transition probability matrix associated with action $a$ and $R_U(s,a)$ the expected reward per second for state $s$ and action $a$ for the uniformized formulation. Alg. \eqref{alg: VI} states how to compute them.
VI consists of computing $u_{n+1}(s)=\underset{a}{\max}\left\{\frac{1}{U}R(s,a) + (P_U(a)u_n)(s)\right\}$
for all states $s$, where $U$ is the uniformization constant associated with $P$ (see \citealp[8.5.1]{putermanMarkovDecisionProcesses2005a} for further details on VI).
According to Puterman's theorem 8.5.6, by stopping VI when $span(u_{n+1}-u_n)<\varepsilon$, we obtain a policy $\pi^{VI}$ with an average reward $\rho^{VI}$ that satisfies:
\begin{equation}
    \label{eq: average reward for VI}
    \rho^{VI} \geq \rho^* - \varepsilon
\end{equation}

From Puterman's section 8.5.4, since any stationary policy is unichain and has an aperiodic transition matrix, the stopping criterion is met after a finite sequence of steps.
According to Theorem 1 of Della Vecchia et al. \citeyearpar{dellavecchiaIllustratedReviewConvergence2012} which is based on Kemp's theorem 3.4.2 \citeyearpar{kempStochasticModellingAnalysis1987}, VI's convergence is geometric.

The adaptation of VI to the admission control problem is presented in Alg. \eqref{alg: VI}.
\begin{center}
\begin{algorithm}
    \caption{VI for Admission Control}
    \begin{algorithmic}
        \STATE \textbf{Improvement}: compute new policy $\pi_{l+1}$ satisfying for all $s=0, \ldots, S-1$:
        \begin{equation*}            
        \forall\, 1\leq i \leq m, \begin{cases}
                r^{(i)}(s)>\nabla u_l(s) &\implies i \in \pi_{l+1}(s)\\
                r^{(i)}(s)<\nabla u_l(s) &\implies i \notin \pi_{l+1}(s)
            \end{cases}
        \end{equation*}
        with $\nabla u_l(s) = u_l(s)-u_l(s+1)$ for $0 \leq s <S$.
        \STATE \textbf{Evaluation}: compute for all $s$:
        \begin{equation*}
            u_{l+1} = R_U^{\pi_{l+1}} + P_U^{\pi_{l+1}} u_l
        \end{equation*}
        where $R_U^{\pi_{l+1}} = \frac{1}{U}R^{\pi_{l+1}}$, $P_U^{\pi_{l+1}} = \frac{1}{U}Z^{\pi_{l+1}}+I$ and $U=\Lambda_{\max} + \mu_{\max}$.
    \end{algorithmic}
    \label{alg: VI}
\end{algorithm}
\end{center}

\section{UCRL for Admission Control}
\label{section: algorithm}
Inspired by UCRL2, we propose the UCRL-AC algorithm (Alg. \ref{alg: UCRL for admission control}) that minimizes the expected regret by optimizing the policy of a CTMDP that is both plausible and optimistic with respect to observed arrival rates.

There are three main steps in the algorithm UCRL-AC.
Step $1$ consists in computing confidence intervals for the global arrival rate $\Lambda$ and for the probability $p^{(i)}$ that an arriving job is of class $i$.
In step $2$, we determine the optimistic CTMDP that is compatible with the confidence intervals computed in step $1$ and update the policy. 
In step $3$, we let the operator apply the policy we updated in step $2$ until the end of the episode.
At the start of a new episode, the estimator of the global arrival rate is reinitialized.
It is updated online at each arrival during the episode. 
We count the number of arrivals of each job classes
and we update the estimator of the probabilities $p^{(i)}$ at the end of the episode.

It is worth noting that
thanks to the structure of the problem, we can easily find the optimistic CTMDP among the plausible ones (Thm \ref{thm: optimistic CTMDP}).
As we do not need to find the optimistic CTMDP while optimizing the policy, 
we can use PI instead of Extended-VI \citep{JMLR:v11:jaksch10a}.
PI produces an exact optimal policy, 
% we use this guarantee to considerably
% reduce the influence of the buffer size $S$ on the regret bound.
and we use this guarantee to eliminate the dependence on 
$S$ from the term that asymptotically dominates over time.
In the following, we describe in detail the three steps of UCRL-AC.

\subsection{Step 1: Confidence Intervals}
Inspired by Gao and Zhou \citeyearpar{gaoLogarithmicRegretBounds2023}
we use the truncated empirical mean estimator of Bubeck et al. \citeyearpar{bubeckBanditsHeavyTail2013}
to construct a confidence interval for the global arrival rate $\Lambda$, and obtain the confidence intervals \eqref{eq: CI}. 

\begin{definition}[Truncated empirical mean \protect{\citep{gaoLogarithmicRegretBounds2023}}]
    \label{def: 3.1}
    The truncated empirical mean estimator for episode $k+1$ 
    is defined as:
    \begin{equation}
        \label{eq: arrival rate estimators}
        \frac{1}{\hat{\Lambda}_{k+1}}=\frac{1}{\nu_k}\sum_{t=1}^{\nu_k} L_t \mathds{1}{\left\{ L_t\leq \sqrt{\frac{2t}{\Lambda_{\min}^2\log\frac{1}{\delta_k}}} \right\}}
    \end{equation}
    where $\nu_k$ is the total number of arrivals within episode $k$, $\delta_k$ controls the precision of the estimator,
    and $L_t$ is an inter-arrival time.
\end{definition}

\begin{proposition}[Confidence intervals]
    \label{proposition: confidence intervals}
    At the start of episode $k\geq 2$,
    with probability at least $1-\delta_{k-1}$, the true global arrival rate lies in the confidence interval defined by
    \begin{equation} \label{eq: CI}
        CI_\Lambda(k)=\Big[\hat{\Lambda}_k - \varepsilon_{k, \Lambda},\hat{\Lambda}_k + \varepsilon_{k, \Lambda}\Big]\cap\Big[\Lambda_{\min},\Lambda_{\max}\Big]
    \end{equation}
    with $\varepsilon_{k,\Lambda}=4\frac{\Lambda_{\max}^2}{\Lambda_{\min}}\sqrt{\frac{2}{\nu_{k-1}}\log\frac{1}{\delta_{k-1}}}$, and $\hat{\Lambda}_k$ defined by \eqref{eq: arrival rate estimators}.

    We note $\hat{p}_{k}^{(i)}$ the empirical estimator of the probability $p^{(i)}$ of an arrival of class $i$ after $k-1$ episodes, that is, $\hat{p}_{k}^{(i)} = N_{k-1}^{(i)}/ N_{k-1}$ where $N_k^{(i)}$ is the total number of arrivals of class $i$ from the first episode to the end of episode $k$ and $N_k$ is the total number of arrivals. 
    With probability at least $1-\delta_{k-1}$, the true distribution belongs to $CI_p(k)$ the set of probability distributions such that for all $p\in CI_p(k)$, 
    \begin{equation}
        \label{eq: l1 inequality for proba}
        \lVert p-\hat{p}_k \rVert_1 \leq \sqrt{\frac{2m}{N_{k-1}}\log \frac{2}{\delta_{k-1}}} = \varepsilon_{k,p}
    \end{equation}
    \end{proposition}
    
Note that since the upper tail of the exponential distribution is not sub-Gaussian,
we use the truncated empirical mean estimator (see \citealp{bubeckBanditsHeavyTail2013}) for which exact bounds are known.
\subsection{Step 2: Optimistic CTMDP and Optimal Policy}
\textbf{Optimistic CTMDP.}
At the end of episode $k$, we determine a confidence interval
for the global arrival rate $\Lambda$ and a confidence set of probability distributions for $p$. 
We note $\mathcal{C}_k$ the confidence set containing the compatible CTMDPs. 
We say that a CTMDP $\mathcal{M}\in \mathcal{C}_k$ if $\mathcal{M}$ corresponds to the global arrival rate $\Lambda$ and class probabilities $p^{(i)}(s)$ such that $\Lambda\in CI_\Lambda(k)$ and $p(s)\in CI_p(k)$ for $s=0, \ldots, S$.

Intuitively, the optimistic CMTDP should put more weight on 
% the 
higher priority classes and less on 
% the
lower priority classes.
Since the holding costs are class-dependent non-decreasing functions of $s$, the priority order of the job classes may change with $s$.
Therefore, the arrival rates of the optimistic CTMDP should be allowed to depend on $s$. 
This is confirmed by the following theorem.
The proof is given in appendix \ref{appendix: B}.

\begin{theorem} \label{thm: optimistic CTMDP}
    Let $\widetilde{\mathcal{M}}$ be the CTMDP in the confidence set $\mathcal{C}$ with the greatest global arrival rate $\tilde{\Lambda}$ and the distributions $\tilde{p}(s)$ over the arrival classes maximizing for each state $s$ the probabilities by priority order according to $r^{(i)}(s)$ given in \eqref{eq: definition r_i(s)}.
    % priority order. 
    $\widetilde{\mathcal{M}}$ achieves the maximal average reward among the CTMDPs in $\mathcal{C}$.
\end{theorem}

\textbf{Search for an optimal policy.}
\label{PI-B}
For PI (Alg. \ref{alg: PI}) and VI (Alg. \ref{alg: VI}), the improvement criterion allows us to choose whether to accept or reject jobs such that $r^{(i)}(s)=\nabla h^\pi(s)$. 
We decide to accept these jobs.
By accepting these jobs, we obtain threshold policies of a certain type: for each state $s$, there is a threshold that gives the rank of the last job class to be accepted when classes are ordered by decreasing expected rewards $r^{(i)}(s), i=1,\ldots,m$.
Thus, the number of policies visited is upper bounded by $(m+1)^S$.
We denote this variant of PI as PI-B.

Alternatively, we can achieve a polynomial complexity by solving a linear program, as initially proposed by Manne \citeyearpar{manneLinearProgrammingSequential1960}.
We recall here 
Puterman's formulation (\citeyear{putermanMarkovDecisionProcesses2005a}, section 8.8):
$$ \begin{aligned} 
&\underset{\rho, \nabla h}{\text{minimize}} \, \rho \qquad\text{ subject to }\forall s, \forall u, \\
&\rho \geq \sum_{i\in u} \lambda^{(i)}(s) \left(r^{(i)}(s)-\nabla h(s)\right) + \mu(s) \nabla h(s-1)
\end{aligned} $$

% Also, when the number of classes $m$ is small, it is tempting to consider Feinberg and Yang parametrization, however 
% MDP optimiste $\lambda^{(i)}(s)$, open question.

\textbf{Remarks}
 \textit{(i)} Feinberg and Yang \citeyearpar{feinbergOptimalityTrunkReservation2011} showed that for state-independent arrival rates, there is an optimal policy with thresholds per class: jobs of a given class are accepted if and only if the the total number of jobs in the queue is smaller than the class threshold. 
When the number of classes $m$ is small, it is tempting to restrict the policy space to such policies.
However, the optimistic CTMDP may have state-dependent arrival rates.
In this case, the optimality of policy with thresholds per class remains an open question.

\textit{(ii)} VI could be combined with Action Elimination \citep[section 8.5.6]{putermanMarkovDecisionProcesses2005a} to guarantee that an optimal policy is reached and ensure the same regret as PI.

\subsection{Step 3: Exploration}
\label{subsection: exploration}
We note $t_k$ the time spent on episode $k$. 
We choose for all $k\geq 1, t_{k+1}=\sum_{p=1}^k t_k$. 
By construction, $t_k = 2^{k-2}t_1$ for $k\geq 2$.
We note $T_k$ the total time spent until the end of episode $k$. For all $k \geq 1, T_k = 2^{k-1}t_1$.

The truncated empirical mean estimator $\hat{\Lambda}_k$ defined by \eqref{eq: arrival rate estimators} can be computed online using the following equation, where 
$\tau$ is the number of job arrivals in episode $k$:
\begin{equation} \label{eq:online_estimate}
    \frac{1}{\hat{\Lambda}_{k,\tau+1}} = \frac{1}{\hat{\Lambda}_{k,\tau}} + \frac{1}{\tau+1}\left(Y_{\tau+1} - \frac{1}{\hat{\Lambda}_{k,\tau}}\right)
\end{equation}
where $Y_\tau = L_\tau \mathds{1}{\left\{ L_\tau \leq \sqrt{\frac{2\tau}{\Lambda_{\min}^2\log\left(\delta_{k-1}^{-1}\right)}} \right\}}$.

The estimator of the global arrival rate is reset at each episode.
Indeed, to bound the regret, we need the sequence $(\delta_k)_{k\geq 1}$ to be decreasing.
However,
since the threshold $2\tau / \left(\Lambda_{\min}^2\log\left(\delta_k^{-1}\right)\right)$ decreases with $\delta_k$, an inter-arrival time that was used in episode $k$ to build the estimator may be discarded in a subsequent episode $k'>k$. 
The reset of the estimator means that the arrivals of previous episodes are not exploited. We could overcome this shortcoming by storing all arrival times, but this would have a memory cost asymptotically proportional to the time horizon.
Despite this limitation, the truncated empirical mean produces confidence intervals that are easier to use than those of the empirical mean.

Unlike classic UCRL algorithms, the exploration time is deterministic for UCRL-AC. 
Indeed, to improve the estimation of an arrival rate, we do not need to visit specific state-action pairs, but to increase the number of arrivals of the corresponding job class.
By specifying a precise exploration time, we can force a high number of arrivals for all classes with high probability.

\begin{algorithm}
    \caption{UCRL-AC}
    \begin{algorithmic}[1]
        \REQUIRE{Rewards $r^{(i)}(s)$,
            lower and upper bounds for the global arrival rate $\Lambda_{\min}$ and $\Lambda_{\max}$,
            time spent in the first episode $t_1$}
        \FOR{episodes $k=1, 2, \ldots$}
        
        \STATE{\textbf{Update}}
        \STATE{1. 
        For episode $1$, set $\Lambda_1=\Lambda_{\max}$ and $p_1=1$.
        For episode $k>1$, compute the confidence intervals for $\Lambda$ (prop. \ref{proposition: confidence intervals}) and $p$
        based on $\hat{\Lambda}_{k-1}$ \eqref{eq: arrival rate estimators} and $\hat{p}_{k-1}$. }
        \STATE{2. Determine the optimistic CTMDP $\widetilde{\mathcal{M}}$ in $\mathcal{C}_k$ (with Theorem. \ref{thm: optimistic CTMDP})
            and compute an associated gain optimal policy $\tilde{\pi}_k$ using for instance PI. Select for the policy $\pi_k=\tilde{\pi}_k$}
       \STATE{\textbf{Exploration}}
        \STATE{3. Accept the jobs according to policy $\pi_k$ for time $t_k$, i.e. until total time $T_k=t_k + T_{k-1}$, with $T_0=0$.
        Compute $\hat{\Lambda}_{k+1}$ online with \eqref{eq:online_estimate} using $\delta_k=1/(\mu t_k)$. When done, compute $\hat{p}_{k+1}^{(i)}=N_k^{(i)}/N_k$}  
        \ENDFOR
    \end{algorithmic}
    \label{alg: UCRL for admission control}
\end{algorithm}

\textbf{Complexity} 
The time and space complexity of UCRL-AC is of same order as those of the policy search algorithm. 
Additional details are provided in appendix \ref{annex: time and space complexity}.

\textbf{Remark} By accepting all customers satisfying $r^{(i)}(s) \geq \nabla h^{\pi_l}(s)$ for all $i=1,\ldots, m$ and $s=0, \ldots, S-1$ like we propose with \hyperref[PI-B]{PI-B}, we obtain at convergence a bias optimal policy, which is a gain optimal policy $\pi$ with maximum bias $h^\pi$ (see appendix \ref{section: VI or PI}). 
Since UCRL-AC relies on a sequence of finite episodes and a bias optimal policy maximizes the short-term return \cite{lewisBiasOptimality2002}, we can hope that a bias optimal policy decreases the expected total regret.

\section{Computing the Bias}
\label{section: computing the bias}
The variant \hyperref[PI-B]{PI-B} of Alg. \eqref{alg: PI} may test up to $(m+1)^S$ different policies in the worst case. Each policy evaluation requires $O(S^3)$ operations to compute the average reward and relative bias for a given policy when using the LAPACK DGESV routine \citep[see section LAPACK Benchmark]{andersonLAPACKUsersGuide1999}. 

It is then relevant to try to reduce the computation cost of each iteration. 
We propose a method to compute the average reward and relative bias of a policy $\pi$ with a linear complexity on the size of the queue $S$. 
Indeed, for the admission control problem and a given policy $\pi$, the long-term average reward $\rho^\pi$ and the relative bias $\nabla h^\pi (s)=h^\pi(s)-h^\pi(s+1)$ for $s \in \{0, 1, \ldots, S-1\}$ have closed-form expressions \eqref{eq: closed-form expression for the average reward} and \eqref{eq: closed-form expression for the relative bias}.

Using the classic result that the average reward $\rho^\pi$ is equal to the scalar product of the invariant distribution with the vector of rewards per second  \citep[eq. (2)]{feinbergOptimalityTrunkReservation2011}, the average reward $\rho^\pi$ has the following closed-form expression

\begin{equation}
    \label{eq: closed-form expression for the average reward}
    \rho^\pi = \frac{\sum_{p=0}^{S}R^\pi(p)\prod_{q=0}^{p-1}\frac{\Lambda^\pi(q)}{\mu(q+1)}}{\sum_{p=0}^{S}\prod_{q=0}^{p-1}\frac{\Lambda^\pi(q)}{\mu(q+1)}}
\end{equation}

\begin{proposition}
    \label{prop: closed-form expression as a matrix product}
    For a given policy $\pi$, the relative bias $\nabla h^\pi$ can be computed by a simple matrix multiplication:
    \begin{equation}
        \label{eq: closed-form expression for the relative bias}
        \nabla H^\pi = U^\pi(\rho^\pi \bm{e}-R^\pi)
    \end{equation}
    where ${\left(\nabla H^\pi\right)}_s = \nabla h^\pi(s) = h^\pi(s)-h^\pi(s+1)$,
    ${\left(R^\pi\right)}_q=R^\pi(q)$, $\bm{e}$ is the column vector composed of ones
    and
    ${\left(U^\pi\right)}_{s,q} = \mathds{1}_{\left\{ q\geq s+1 \right\}}\frac{1}{\mu(s+1)}\prod_{p=s+1}^{q-1} \frac{\Lambda^\pi(p)}{\mu(p+1)}$.
\end{proposition}

\textbf{Remark} There is an alternative to the computation of the matrix $U^\pi$ to obtain $\nabla h^\pi$.
Once the average reward $\rho^\pi$ is known, 
we can compute the relative bias of one of the two states $s=0$ and $s=S-1$ by applying \eqref{eq: average reward for admission control} to $s=0$ and $s=S$:
\begin{equation*}
    \rho^\pi = R^\pi(0)-\Lambda^\pi(0) \nabla h^\pi(0) = \mu(S)\nabla h^\pi(S-1)
\end{equation*}
We can achieve a linear complexity for the computation of $\nabla h^\pi$ by using equation \eqref{eq: average reward for admission control} recursively,
starting with the relative bias of one of these two states.
Knowing $\pi$, $\rho^\pi$, the service rate $\mu(s)$ and the expected rewards $r^{(i)}(s)$, we can deduce from $\nabla h^\pi(s-1)$ the value of $\nabla h^\pi(s)$ and reciprocally.
The improvement step requires to compare $r^{(i)}(s)$ and $\nabla h^\pi (s)$ for all $s$.
Its complexity is $O(mS)$.

\section{Regret Analysis}
\label{section: regret}
Remember that we study the regret to measure the effectiveness of our algorithm over the entire learning process, and not just to measure the performance of a final policy.
We note $\Delta_k$ the regret for episode $k$ of length $t_k$
in which policy $\pi_k$ is used:
\begin{equation*}
    \Delta_k = t_k \rho^* - \sum_{s,i}\nu_k(s,i)r^{(i)}(s)
\end{equation*}
where $\nu_k(s,i)$ is the number of job of class $i$ that are accepted while the system is in state $s$ during episode $k$.
We note
$\Delta_k^{in}$ the regret for episode $k$
when the true CTMDP lies in the confidence set $\mathcal{M} \in \mathcal{C}_k$ and
$\Delta_k^{out}$ the regret for episode $k$
when the true CTMDP is not in the confidence set:
\begin{equation*}
    \Delta_k^{in}  = \Delta_k \mathds{1}_{\mathcal{M}\in\mathcal{C}_k}, \quad
    \Delta_k^{out} = \Delta_k \mathds{1}_{\mathcal{M}\notin\mathcal{C}_k}
\end{equation*}

When the true CTMDP is in the confidence set,
the number of arrivals may be insufficient
to effectively bound the regret.
Thus, we decompose $\Delta_k^{in}$ into two terms:
$\Delta_k^{good}$, which corresponds to the case where enough jobs have appeared,
and $\Delta_k^{bad}$ which corresponds to the case where too few jobs have arrived. 
By noting $\nu_k$ the number of arrivals in episode $k$, and
with $\Delta_{k+1}^{bad} = \Delta_{k+1}^{in}\mathds{1}_{\nu_k< \Lambda t_k/2}$ and $\Delta_{k+1}^{good} = \Delta_{k+1}^{in}\mathds{1}_{\nu_k \geq \Lambda t_k/2}$, the regret can be written as:
\begin{equation*}
    \Delta_k = \Delta_k^{out}+\Delta_k^{bad}+\Delta_k^{good}
\end{equation*}
and the expected regret $\mathcal{R}_{(K)}$ for episodes $1$ to $K$ is $\mathcal{R}_{(K)}=\E{\sum_{k=1}^K \Delta_k}$, with
\begin{equation*}
    \mathcal{R}_{(K)} = \mathcal{R}_{(K)}^{out} +\mathcal{R}_{(K)}^{bad} + \mathcal{R}_{(K)}^{good}
\end{equation*}

With class-dependent holding costs, the reward order can change, hence the optimistic CTMDP may have arrival rates that depend on the state of the system. 
To effectively bound the regret in this case, in Theorem \ref{theorem: bound for relative bias}, we extend the results of Feinberg and Yang \citeyearpar{feinbergOptimalityTrunkReservation2011} to state-dependant arrival rates.
\begin{theorem}
    \label{theorem: bound for relative bias}
    We suppose state-dependent arrival rates $\lambda^{(i)}(s)$ for $s=0,\ldots, S$ and  $i=1, \ldots, m$ associated to a CTMDP $\mathcal{M}$ in the confidence set $\mathcal{C}$. For any policy $\pi$, we note $\rho^{\pi}$ its average reward and $h^\pi$ its bias in $\mathcal{M}$.
    We have:
    
    (i) $\rho^\pi \leq \sum_i \lambda^{(i)}(0) R^{(i)} \leq \Lambda_{\max} R_{\max}$
        
    (ii) If $\pi$ is gain optimal, $0 < \nabla h^\pi(s) \leq \frac{\rho^\pi}{\mu_{\max}}\leq \frac{\Lambda_{\max} R_{\max}}{\mu_{\max}}$.
\end{theorem}

While PI yields a gain optimal policy, its convergence speed is unknown. VI converges geometrically fast, but not necessarily to a gain optimal policy due to the stopping criterion. 
 Thus, Theorem \ref{theorem: bound for relative bias} can be applied to PI, but not to VI.
 A more thorough explanation can be found in appendix \ref{section: VI or PI}.
 
The regret bound for UCRL-AC is given in Theorem \ref{thm: upper bound of the expected regret}:

\begin{theorem}
    \label{thm: upper bound of the expected regret}
    For a fixed global arrival rate $\Lambda = \sum_i \lambda^{(i)}$, if the immediate rewards $R_1,\ldots, R_m$ are bounded, then the expected regret over $K$ episodes $\mathcal{R}_{(K)}$ is upper bounded. 
    For Policy Iteration,
    \begin{equation}
    \label{eq: regret for policy iteration}
        \mathcal{R}_{(K)} \leq a \sqrt{T_K \log (2\mu T_K)} + b \left(1+\log_2\left(\frac{T_K}{t_1}\right)\right) + c
    \end{equation}
    with 
    \begin{align} 
    a &= 14\left(4\frac{\Lambda_{\max}^2}{\Lambda_{\min}\sqrt{\Lambda}}+ \sqrt{m\Lambda}\right)\left(1+\frac{\Lambda_{\max}}{\mu_{\max}}\right)R_{\max} \\
    b &= \left(\frac{4}{\mu}+\frac{14}{\Lambda}\right)\rho^* + \frac{S\Lambda_{\max}R_{\max}}{\mu_{\max}}\ 
    \\ c &= \rho^* t_1 
     \end{align} 

For Value Iteration,
\begin{equation}
    \label{eq: regret for value iteration}
        \mathcal{R}_{(K)} \leq a' \sqrt{T_K \log (2\mu T_K)} + b' \left(1+\log_2\left(\frac{T_K}{t_1}\right)\right) + c'
\end{equation}
with 
    \begin{align} 
    a' &= 14\left(4\frac{\Lambda_{\max}^2}{\Lambda_{\min}\sqrt{\Lambda}}+ \sqrt{m\Lambda}\right)\left(R_{\max} + V \right) \\
    b' &= \left(\frac{4}{\mu}+\frac{14}{\Lambda}\right)\rho^* +R_{\max}+V 
    \\ c' &= \rho^* t_1 
     \end{align} 
 and
    $V = \underset{s}{\max}\left\{\frac{\Lambda_{\max} R_{\max}}{\mu} \sum_{q=s+1}^S \prod_{p=s+1}^{q-1} \frac{\tilde{\Lambda}^\pi(p)}{\mu(p+1)}\right\}$
\end{theorem} 

\begin{figure*}[t]
    \captionsetup[subfigure]{}
    % \hspace{-0.6cm}
    \subcaptionbox{$S=20, \mu=0.3$\vspace{0.5cm}}{
        \includegraphics{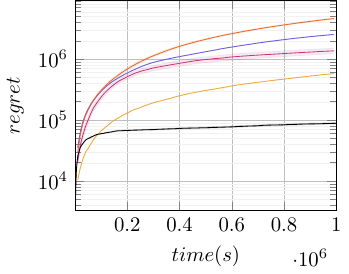}
        }%
    \hspace{-2.5cm}
    \subcaptionbox{$S=20, \mu=0.4$}{ 
        \includegraphics{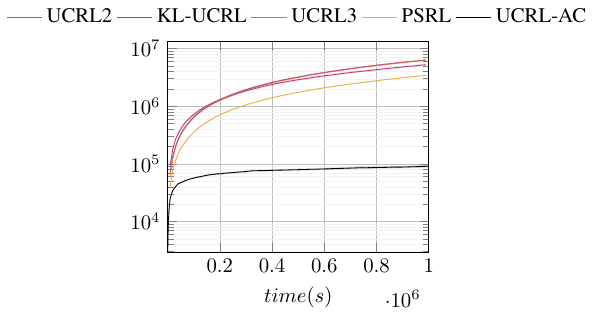}
        }%
    \hspace{-3cm}
    \subcaptionbox{$S=20, \mu=0.5$}{
        \includegraphics{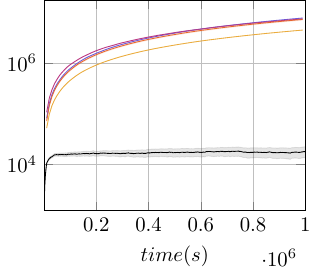}
    }
    \subcaptionbox{$S=50, \mu=0.3$         }{
         \includegraphics{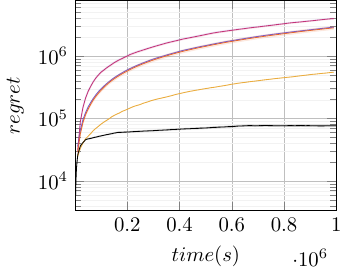}
        }%
    \hspace{-0.1cm}
    \subcaptionbox{$S=50, \mu=0.4$}{ 
        \includegraphics{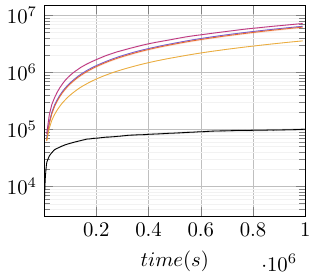}
    }
    \hspace{-0.1cm}
    \subcaptionbox{$S=50, \mu=0.5$}{
        \includegraphics{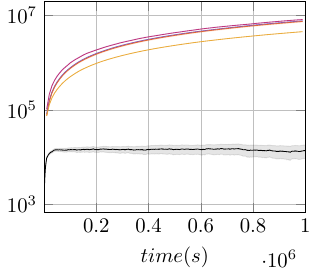}
    }
\caption{Comparison of UCRL-AC with UCRL2, PSRL, KL-UCRL and UCRL3 with buffers of size $20, 50$ and individual service rates equal to $0.3, 0.4$ and $0.5$. We consider $5$ servers, $2$ job classes with immediate rewards $R_1=20$ and $R_2=10$ and arrival rates $\lambda_1 = 1$ and $\lambda_2=1$ respectively, and holding cost $C(t)=0.1t$ for both classes. For UCRL-AC, we used $\Lambda_{\min}=1$ and $\Lambda_{\max}=4$.}
\label{fig: empirical comparison}
\end{figure*}

In the infinite server case ($c=S$), $S/\mu_{\max}=1/c$ and the influence of $S$ disappears.
We provide an intuition:
when $c$ is equal to $S$ and $S$ grows to $\infty$, 
the queue tends to behave as an $M/M/\infty$.
In this limit, all jobs are accepted and the dependence on $S$ naturally disappears.
In fact, when $c=S$, the corresponding term in the regret is proportional to $\rho^*$,
which can still grow with $S$, but without exceeding the upper bound $\rho_{\max}$.

In comparison, applying Zhang and Ji's algorithm \citeyearpar{zhangRegretMinimizationReinforcement2019a} leads to an upper bound on the regret of $\tilde{O}\left( \sqrt{mS^2T/\mu(S)} \right)$. Indeed, we can use their algorithm since the span $H$ of the bias function of an optimal policy can be bounded by applying Theorem \ref{theorem: bound for relative bias}: $H = h^*(0)-h^*(S)\leq \Lambda_{\max}R_{\max} S/\mu(S)$. With their approach, the term that dominates asymptotically with time still depends on the queue size $S$.
% Then, we can obtain  an  by applying Zhang and Ji's algorithm \citeyearpar{zhangRegretMinimizationReinforcement2019a}:
% the
% bound is linear in $S$ for a finite number of servers $c<S$ and linear in $\sqrt{S}$ for the infinite server case $c=S$. 
% Yet, $S$ appears in the term that dominates asymptotically with time. 
% In this article, we reduce the influence of $S$ by applying techniques inspired by UCRL2's original proof.

\section{Experiments}
\label{section: experiments}
We slightly modify Alg. \ref{alg: UCRL for admission control} to further reduce the empirical regret by learning an estimate of $\Lambda_{\max}$. 
The theoretical bound still holds (see appendix \ref{appendix: empirical regret}):
\begin{proposition}
    \label{prop: learning lambda_max and lambda_min}
    Let $\hat{\Lambda}_k$ be the arrival rate estimator defined in \eqref{eq: arrival rate estimators}. 
    Let $\varepsilon_{k+1}=\frac{4}{\Lambda_{\min}}\sqrt{\frac{2}{\nu_{k}}\log\frac{1}{\delta_{k}}}$d, $\overline{\Lambda}_{1}=\Lambda_{\max}$ and, for $k\geq 1$,
    \begin{equation}
        \overline{\Lambda}_{k+1} = \begin{cases}
            \min(\Lambda_{\max}, \hat{\Lambda}_k+\Lambda_{\max}^2 \varepsilon_k) & \hat{\Lambda}_k\varepsilon_k \geq 1 \\
            \min\left(\Lambda_{\max}, \frac{\hat{\Lambda}_k}{1-\hat{\Lambda}_k \varepsilon_k}, \hat{\Lambda}_k+\Lambda_{\max}^2 \varepsilon_k\right) & \text{otherwise}
            
        \end{cases}
    \end{equation}
    The regret bounds in Theorem \ref{thm: upper bound of the expected regret} still hold if we use $\overline{\Lambda}_k$ in episode $k$ instead of $\Lambda_{\max}$.
\end{proposition}

We compare the performance of our algorithm with UCRL2,  PSRL \citep{osbandMoreEfficientReinforcement2013}, KL-UCRL \citep{talebiVarianceAwareRegretBounds2018b} and UCRL3 \citep{bourelTighteningExplorationUpper2020a} using the code proposed by the authors of UCRL3. 
For PSRL, we used a beta prior on the normalized reward. 
For each algorithm, the action space is encoded as "selecting the $k$ most priority job classes", where $k$ is controlled.

We focus our experiments on the influence of the system's load and the size of the queue on the regret.
The system is said overloaded (resp. under-loaded) if the global arrival rate is bigger (resp. smaller) than the maximal service rate. 
We present results for the overloaded regime, the regime in which the global arrival rate is equal to the maximal service rate, and the under-loaded regime.
In our experiments, there are two job classes, both with an arrival rate of $1$. The system is of size $S$ in $\{20, 50\}$, has $5$ identical servers of service rate $\mu$ in $\{0.3, 0.4, 0.5\}$. The maximal service rate is then either $1.5$, $2$ or $2.5$ while the global arrival rate is $2$. As PI and VI lead to exactly the same regrets, we only plot the results for PI.

Fig. \ref{fig: empirical comparison} compares the regrets obtained with UCRL2, PSRL, KL-UCRL, UCRL3 and UCRL-AC over $100$ trajectories. 
Thick lines represent the average trajectory, while colored areas represent the 95\% confidence interval.

For $S=20$ and $\mu=0.3$, we observe that for $S=20$ and $\mu=0.3$, PSRL has smaller regrets than UCRL-AC at the beginning. 
We believe that this is because UCRL-AC suffers from a large overestimation of the global arrival rate, which 
leads to too cautious policies.
% initially leads to cautious policies favoring the highest priority class too much.
% a slow start of UCRL-AC compared to PSRL. 
Otherwise, UCRL-AC achieves lower empirical regrets than the other algorithms, especially as the size of the queue $S$ increases.

\begin{figure}
    \includegraphics{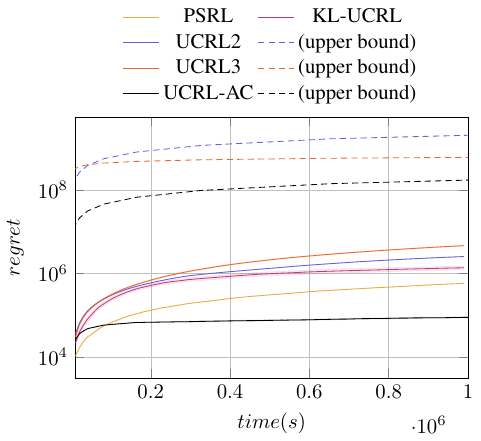}
    \caption{Theoretical upper bounds of UCRL2, UCRL3 and UCRL-AC and empirical regrets of UCRL2, PSRL, KL-UCRL, UCRL3 and UCRL-AC with the setting of Fig. \ref{fig: empirical comparison}(a).}
\vspace{-0.5cm}
\label{fig: upper bounds}
\end{figure}

We compare in the next experiment the theoretical upper bounds.
% of UCRL2, UCRL3 and UCRL-AC.
For that, we need to calculate the diameter of the MDP $D$.
We discuss in the appendix \ref{appendix: diameter lower bound} how to compute it.
Fig. \ref{fig: upper bounds} presents 
an example of the upper bounds of our algortihm, UCRL2 and UCRL3.
% the theoretical bounds and empirical trajectories for $S=20$, $\mu=0.3$.
% We see that our upper bound is smaller than the other two, but still very loose as all algorithms achieve far smaller regrets.
We find that our upper bound is an order of magnitude lower than UCRL2 and UCRL3 and that, for all algorithms, the empirical regrets are far smaller than the theoretical guarantees.

The upper bounds of KL-UCRL and PSRL are not shown here, since our problem does not satisfy one of the assumptions for KL-UCRL to derive an upper bound (see \citealp{talebiVarianceAwareRegretBounds2018b}, Remark 17: the transition probability matrix of an optimal policy is not necessarily contracting for the span), and we have not found theoretical upper bounds for PSRL for the infinite horizon \citep{osbandPosteriorSamplingReinforcement2016a}.

Our code is available at: \url{https://github.com/luweber21/ucrl-ac}.

\section{Conclusion and Future Work}
\label{section: conclusion}
In the general case for undiscounted reinforcement learning,
the lower bound on the expected regret for any learning algorithm is $\Omega(\sqrt{DXAT})$. 
We propose an algorithm for the admission control problem that achieves a smaller regret with no dependence on the diameter $D$. 
In the infinite server case,
the dependence on the buffer size $S=X-1$ completely vanishes.
For the regime in which the time $T$ dominates,
the leading term does depend on $S$ and
our bound becomes $O(\sqrt{mT\log T})$.

We conjecture that with some adaptations, our approach can be extended to 
% {\color{red}state-dependent arrival rates that are constant for $s<c$ and non-increasing otherwise} 
non-increasing state-dependent arrival rates. To that end, 
classical stochastic length of episodes should be used: the episode changes
every time the number of visits to a state doubles. However, this would multiply the $\sqrt{T\log T }$ term of the regret bound by $\sqrt{S}$.

Finally, we think it would be valuable to extend this work to scenarios where additional parameters, such as the service rate, are unknown.

\section*{Acknowledgements}
This work was supported by the Direction Générale de l’Armement.

\section*{Impact Statement}
This paper presents work whose goal is to advance the field of Machine Learning. There are many potential societal consequences of our work, none which we feel must be specifically highlighted here.

\bibliography{biblio.bib}
\bibliographystyle{icml2024}

%%%%%%%%%%%%%%%%%%%%%%%%%%%%%%%%%%%%%%%%%%%%%%%%%%%%%%%%%%%%%%%%%%%%%%%%%%%%%%%
%%%%%%%%%%%%%%%%%%%%%%%%%%%%%%%%%%%%%%%%%%%%%%%%%%%%%%%%%%%%%%%%%%%%%%%%%%%%%%%
% APPENDIX
%%%%%%%%%%%%%%%%%%%%%%%%%%%%%%%%%%%%%%%%%%%%%%%%%%%%%%%%%%%%%%%%%%%%%%%%%%%%%%%
%%%%%%%%%%%%%%%%%%%%%%%%%%%%%%%%%%%%%%%%%%%%%%%%%%%%%%%%%%%%%%%%%%%%%%%%%%%%%%%
\newpage
\appendix
\onecolumn

\appendix
\section{Lower Bound of the Diameter}
\label{appendix: diameter lower bound}
The diameter $D$ is defined by Jaksch et al. as 
\begin{equation}
    \label{eq: diameter}
    D = \underset{s\neq s'}{\max}\;\underset{\pi}{\min}\;\E{T(s'|\pi, s)}
\end{equation}
where $T(s'|\pi, s)$ is the random variable for the first time step in which state $s'$ is reached, starting in initial state $s$ and following policy $\pi$. 
Since we are dealing with a birth-and-death process, the diameter either corresponds to $\underset{\pi}{\min}\;\E{T(S|\pi, 0)}$ or $\underset{\pi}{\min}\;\E{T(0|\pi, S)}$. The best policy to go from $0$ to $S$ is to accept all jobs, while the best policy to go from $S$ to $0$ is to reject all of them. We note $\pi_\circ$ and $\pi_\times$ these two policies respectively.

% $\E{T(S|\pi_\circ, 0)}$ and $\E{T(0|\pi_\times, S)}$ can be computed using first step analysis (see chapter 5 of Privault's book \citeyearpar{privaultUnderstandingMarkovChains2018}). 
% Let us note $\tau_s$ the expected number of steps to reach state $0$ starting in state $s$. Then, $\tau_s=1 + P^{\pi_\times}(s-1|s)\tau_{s-1} + P^{\pi_\times}(s|s)\tau_{s} + P^{\pi_\times}(s+1|s)\tau_{s+1}$, with $P^{\pi_\times}$ the transition probability matrix. 
% We are dealing with a CTMDP, so $P^{\pi_\times}$ actually derives from the infinitesimal generator $Z^{\pi_\times}$ and depends on the uniformization constant, which implies that the diameter depends on this constant as well. We note $U$ this constant.
% Since $(1-P^{\pi_{\times}}(s|s)) \tau_s \geq 1 +  P^{\pi_{\times}}(s-1|s) \tau_{s-1}$, we can show by recurrence on $s$ that:
% \begin{equation}
%     \tau_S \geq \frac{\prod_{s=2}^S P^{\pi_{\times}}(s-1|s)}{\prod_{s=2}^S \left(1-P^{\pi_{\times}}(s|s)\right)}\tau_{1} \geq \frac{\prod_{s=2}^S P^{\pi_{\times}}(s-1|s)}{\prod_{s=2}^S \left(1-P^{\pi_{\times}}(s|s)\right)}
% \end{equation}

We compute a lower bound of the diameter following the example of Anselmi et al.
The diameter is lower bounded by the time to go from state 0 to state $S$ with policy $\pi_{\circ}$. 
We note $\tau^\pi(s_1, s_2)$ the expected time to go from state $s_1$ to state $s_2$ with policy $\pi$. 
Since the diameter is defined for an MDP, we uniformize the CTMDP.
We choose the normalization constant $U=\Lambda+\mu(S)$.

\begin{equation*}
    D \geq \tau^{\pi_{\circ}}(0, S) \geq \tau^{\pi_{\circ}}(S-1, S)
\end{equation*}

In $S-1$, the system goes to state $S$ with probability $\Lambda / U$ and the time to reach $S$ is equal to $1$.
Otherwise, with probability $1-\Lambda / U$, the system goes to state $S-2$ or stays in state $S-1$.
In this case, the time to move from $S-1$ to $S$ is lower bounded by the return time to $S-1$ for a chain truncated at $S-1$.
% This return time is in turn lower bounded by the return time to $S-1$ in the real chain.
Finally, the return time is the inverse of the stationary measure:

\begin{align*}
    D &\geq (1-\Lambda/U) \left(\prod_{x=1}^{S-1}\frac{\mu(x)}{\Lambda}\right)\sum_{s=0}^{S-1} \prod_{x=1}^{s}\frac{\Lambda}{\mu(x)} \\
    &\geq \left(1-\frac{\Lambda}{U}\right)\sum_{s=0}^{S-1}\prod_{x=s+1}^{S-1}\frac{\mu(x)}{\Lambda} 
\end{align*}

We highlight three cases: (i) the $M/M/1/S$, (ii) $M/M/S/S$ queues, (iii) the general case of a $M/M/c/S$ queue with $c<S$.
\begin{enumerate}
    \item For a $M/M/1/S$ queue and $\mu \neq \Lambda$, we have 
    \begin{equation*}
        D \geq \frac{\mu}{\Lambda + \mu} \frac{(\mu/\Lambda)^S-1}{\mu/\Lambda - 1}
    \end{equation*}

    \item For a $M/M/S/S$ queue, 
    \begin{equation*}
        D \geq \frac{S\mu}{\Lambda + S\mu} \left(\frac{\mu}{\Lambda}\right)^{S-1} \left(S-1\right)! \sum_{s=0}^{S-1}\frac{1}{s!}\left(\frac{\Lambda}{\mu}\right)^s 
    \end{equation*}
    \item Finally, the formula for a $M/M/c/S$ queue with $c<S$ and $c\mu\neq \Lambda$ is:
    \begin{equation*}
        D \geq \frac{c\mu}{\Lambda + c\mu} \left[ 
        \frac{\left(c\mu/\Lambda\right)^{S-c}-1}{c\mu/\Lambda-1} + c^{S-c-1}c! \left(\frac{\mu}{\Lambda}\right)^{S-1} \sum_{s=0}^{c-1}\frac{1}{s!}\left(\frac{\Lambda}{\mu}\right)^s
        \right]
    \end{equation*}
\end{enumerate}

\section{Confidence Intervals for the Arrival Rate and the Class Probabilities}
\label{appendix: A}
\paragraph{Proof of proposition \ref{proposition: confidence intervals}}
Equation \eqref{eq: l1 inequality for proba} results from
Weissman et al. \citeyearpar{weissmanInequalitiesL1Deviation2003}, Theorem 2.1.

We now prove equation \eqref{eq: CI}.
Let us note $L$ the random inter-arrival time between two arrivals. 
With $\Lambda_{\min} \leq \Lambda$, $\E{|L|^2}=\frac{2}{\Lambda^2}\leq \frac{2}{\Lambda_{\min}^2}$.
From lemma 1 of Bubeck et al. \citeyearpar{bubeckBanditsHeavyTail2013}\footnote[1]
{The demonstration that appears in the article proves the upper bound for $1/\Lambda-1/\hat{\Lambda}_k$,
    but the same steps can be followed for $1/\hat{\Lambda}_k-1/\Lambda$.}, with $\hat{\Lambda}_{k}$ defined by \eqref{eq: arrival rate estimators}, we know that the following inequality holds with probability at least $1-\delta_k$:
\begin{equation*}
    \left| \frac{1}{\Lambda}-\frac{1}{\hat{\Lambda}_{k}} \right| \leq \frac{4}{\Lambda_{\min}}\sqrt{\frac{2\log(\delta_{k-1}^{-1})}{\nu_{k-1}}}
\end{equation*}
The difference $\left|\Lambda - \hat{\Lambda}_k\right|$ can be bounded using the fact that $\Lambda_{\max}$ is greater than $\Lambda$ and $\hat{\Lambda}_k$: 
\begin{equation*}
\left| \Lambda-\hat{\Lambda}_{k} \right| \leq \frac{\Lambda_{\max}^2}{\Lambda\hat{\Lambda}_{k}}\left| \Lambda-\hat{\Lambda}_{k} \right| = \Lambda_{\max}^2 \left| \frac{1}{\Lambda}-\frac{1}{\hat{\Lambda}_{k}} \right|
\end{equation*}
Therefore, with probability at least $1-\delta_k$:
\begin{equation*}
    \Lambda \in \left[ \hat{\Lambda}_k-\varepsilon_k, \hat{\Lambda}_k+\varepsilon_k \right]
\end{equation*}
where 
\begin{equation*}
    \varepsilon_k=4\frac{\Lambda_{\max}^2}{\Lambda_{\min}}\sqrt{\frac{2}{\nu_k}\log \frac{1}{\delta_k}}
\end{equation*}

\section{Optimistic CTMDP}
\label{appendix: B}
\paragraph{Proof of theorem \ref{thm: optimistic CTMDP}}
We have two results to prove:
\begin{enumerate}
    \item for a fixed distribution on the classes $p$, the CTMDP with the higher global arrival rate achieves higher average rewards;
    \item for a fixed global arrival rate, the CTMDP which puts as much weights as possible on the higher priority jobs achieves higher average rewards. The job distributions are allowed to depend on the state of the system.
\end{enumerate}

\textit{For $1$:}
Let $M$ be a CTMDP with global arrival rate $\Lambda$ and distribution $p$ over the job types. Let $M'$ be a CTMDP with a global arrival rate $\Lambda'\geq\Lambda$ and the same distribution $p$ over the job classes.
Let us consider a stochastic policy $\pi$ such that in state $s$, a customer of type $i$ is accepted with probability $\pi(i|s)$. 
We define another policy $\pi'$ such that $\pi'(i|s)=\frac{\Lambda}{\Lambda'}\pi(i|s)$. Then, $\pi$ applied to $M$ and $\pi'$ applied to $M'$ lead to the same Markov chain and therefore the same average reward.
Thus, for any policy on $M$, there is a policy on $M'$ achieving the same average reward. Hence, the optimal average reward achieved by $M'$ is at least as big as the optimal average reward achieved by $M$.

\textit{For $2$:}
Let $p(s)$ be a distribution over the job classes for state $s$ and $\pi$ be a stochastic policy. Let $\tilde{p}(s)$ be the distribution over the arrival types in the confidence set maximizing the probabilities by order of priority for state $s$. 
Let us note $k$ an integer such that $\forall j\leq k, \tilde{p}^{(j)}(s) \geq p^{(j)}(s)$ and $\forall j > k, \tilde{p}^{(j)}(s) \leq p^{(j)}(s)$. 
We define $\tilde{\pi}$ by $\tilde{\pi}(j|s)=\pi(j|s)$ if $j>k$ and $\tilde{\pi}(j|s)=\frac{p^{(j)}(s)}{\tilde{p}^{(j)}(s)}\pi(j|s)+\left( 1-\frac{p^{(j)}(s)}{\tilde{p}^{(j)}(s)} \right)\sum_{y > k} \frac{p^{(y)}(s)-\tilde{p}^{(y)}(s)}{\sum_{t>k} p^{(t)}(s)-\tilde{p}^{(t)}(s)}\pi(y|s)$ if $j \leq k$.

We note $Z_p^\pi$ the infinitesimal associated with policy $\pi$ and arrival distribution $p(s), s=0, \ldots, S$. We have 
\begin{align*}
    Z_{\tilde{p}}^{\tilde{\pi}}(s, s+1) &= \sum_i \Lambda \tilde{p}^{(i)}(s) \tilde{\pi}(i|s) \\
    &= \Lambda \sum_{i\leq k} \left( p^{(i)}(s) \pi(i|s) + \left(\tilde{p}^{(i)}(s)-p^{(i)}(s)\right)\sum_{j>k} \frac{p^{(j)}(s)-\tilde{p}^{(j)}(s)}{\sum_{t>k}p^{(t)}(s)-\tilde{p}^{(t)}(s)}\pi(j|s)  \right) + \sum_{i > k} \tilde{p}^{(i)}(s) \pi(i|s) \\
    &= \Lambda \sum_{i\leq k} p^{(i)}(s) \pi(i|s) + \Lambda \sum_{i > k} \left( p^{(i)}(s)-\tilde{p}^{(i)}(s) \right)\pi(i|s) + \Lambda \sum_{i>k} p^{(i)}(s)\pi(i|s) \\
    &= \Lambda \sum_i p^{(i)}(s) \pi(i|s) \\
    &= Z_p^{\pi}(s, s+1)
\end{align*}
where we used for the third equality that $\sum_{t>k} p^{(t)}(s)-\tilde{p}^{(t)}(s) = \sum_{t\leq k} \tilde{p}^{(t)}(s)-p^{(t)}(s)$.

Thus, a CTMDP $M'$ with policy $\tilde{\pi}$ and distribution $\tilde{p}$ over the arrivals is described by the same infinitesimal generator as a CTMDP $M$ with policy $\pi$ and distribution $p$ over the arrivals. Therefore, they have the same stationary distribution.

Let us compare the corresponding average rewards.
We note $R_p^\pi(s)$ the expected reward per second in state $s$ with policy $\pi$ and distribution $p$ over the arrival types.
\begin{align*}
    \begin{split}
    R_{\tilde{p}}^{\tilde{\pi}}(s) &= \Lambda \sum_{i\leq k} \left( p^{(i)}(s) \pi(i|s) + \left(\tilde{p}^{(i)}(s)-p^{(i)}(s)\right) \sum_{j>k} \frac{p^{(j)}(s)-\tilde{p}^{(j)}(s)}{\sum_{t>k}p^{(t)}(s)-\tilde{p}^{(t)}(s)}\pi(j|s)\right)r^{(i)}(s) \\&\qquad + \Lambda\sum_{i > k} \tilde{p}^{(i)}(s) \pi(i|s)r^{(i)}(s)
    \end{split}
    \\
    \begin{split}
    &\geq \Lambda \sum_{i\leq k} p^{(i)}(s) \pi(i|s)r^{(i)}(s) + \Lambda \sum_{i\leq k}\left(\tilde{p}^{(i)}(s)-p^{(i)}(s)\right)\sum_{j>k} \frac{p^{(j)}(s)-\tilde{p}^{(j)}(s)}{\sum_{t>k}p^{(t)}(s)-\tilde{p}^{(t)}(s)}\pi(j|s)r^{(j)}(s) \\&\qquad + \Lambda\sum_{i > k} \tilde{p}^{(i)}(s) \pi(i|s)r^{(i)}(s)
    \end{split}
    \\
    &= \Lambda \sum_{i\leq k} p^{(i)}(s) \pi(i|s) r^{(i)}(s) + \Lambda \sum_{i>k} \left( p^{(i)}(s) - \tilde{p}^{(i)}(s) \right) \pi(i|s) r^{(i)}(s) + \Lambda \sum_{i>k}\tilde{p}^{(i)}(s) \pi(i|s)r^{(i)}(s) \\
    &= \Lambda \sum_i p^{(i)}(s) \pi(i|s) r^{(i)}(s) \\
    &= R_{p}^\pi(s)
\end{align*}
Hence, $R_{\tilde{p}}^{\tilde{\pi}}(s) \geq R_p^\pi(s)$.

Combining the two previous results, both CTMDPs have the same stationary distribution but $M'$ yields bigger rewards. Therefore, $M'$ achieves bigger average rewards than $M$.

\section{Validity of Feinberg and Yang's Results for State-Dependent Arrival Rates}
\label{appendix: C}
\textbf{Proof of Theorem \ref{theorem: bound for relative bias}}
We want to prove that all the preliminary results of Feinberg and Yang we need for our proofs remain for state-dependent arrival rates.
We suppose that for all $s=0, \ldots, S-1$, $\sum_{i=1}^m \lambda^{(i)}(s)=\Lambda$.

Most of the proofs do not need any modification, except that all coefficients $\lambda^{(j)}$ ($\lambda^{(j)}$ in their notation) are replaced with $\lambda^{(j)}(s)$. We only list those that require changes.

\begin{itemize}
    \item Theorem 3.1: This theorem states that an optimal policy will not refuse all job classes if a server is free.
    The proof is based on a comparison with an M/M/c/loss queue for which jobs are discarded if no server is free at the arrival time.
    Once we remark that for $s<c, \lambda^{(i)}(s)=\lambda^{(i)}(0)$ for all $i=1, \ldots, m$, the comparison holds.
    
    \item Theorem 3.2: At some point, the relative bias $\nabla h^\pi$ of some particular state $s^\pi$ is compared with the maximal immediate expected reward $r^{(i)}(s^\pi)$. 
    If the arrival rates depend on the state, then it is possible for some to be equal to zero, and the max shoud be taken among the immediate expected reward with arrival rates strictly bigger than zero.
    However, observe that $\underset{j=1, \ldots, m}{\max}\, r^{(j)}(s)=\underset{j=1, \ldots, m}{\max}\left\{ r^{(j)}(s), \lambda^{(j)}(s) > 0 \right\}$ since $\lambda^{(j)}(s)=0$ can only be true for the lowest-priority classes.
    
    \item Lemma 3.3: The proof requires that for any state $s$ and policy $\pi$, the expected reward per second $R^\pi(s)$ is smaller than $\tilde{R}=\sum_i \lambda^{(i)}(0) r^{(i)}(0)$.
    We add at the beginning of the proof that the optimistic CTMDP is such that $p^{(i)}(s), i=1, \ldots, m$ maximizes $\sum_{i}p^{(i)}(s) r^{(i)}(s)$.
    Thus, 
    \begin{align*}
        R^\pi(s) 
            &=\Lambda \sum_{i\in\pi(s)} p^{(i)}(s)r^{(i)}(s) &&\text{by definition}\\
            &\leq\Lambda \sum_{i\in\pi(s)} p^{(i)}(s)\left[r^{(i)}(s)\right]^+ \\
            &\leq \Lambda \sum_{i\in\pi(s)} p^{(i)}(s)\left[r^{(i)}(0)\right]^+ &&\text{since $r^{(i)}(s)$ is non-increasing} \\
            &= \Lambda \sum_{i} p^{(i)}(s)r^{(i)}(0) &&\text{because $r^{(i)}(0)=R^{(i)}>0$} \\
            &\leq \Lambda \sum_{i} p^{(i)}(0)r^{(i)}(0) &&\text{as the vector $p(0)$ maximizes $\sum_i p^{(i)}(0)r^{(i)}(0)$} \\
            &= \tilde{R}
    \end{align*}
    
    \item Lemma 3.7: the original proof obtains at some point the inequality $\sum_{j=1}^m \lambda^{(j)}(s+1)\left[ r^{(j)}(s+1)-\nabla h^\pi(s+1) \right]^+ < \sum_{j=1}^m \lambda^{(j)}(s+2)\left[ r^{(j)}(s+2)-\nabla h^\pi(s+2) \right]^+$ to deduce that $\nabla h^\pi(s+1) > \nabla h^\pi(s+2)$. 
    This deduction is less obvious when the arrival rates depend on the state of the system $s$.
    To prove it, recall that $\Lambda = \sum_{j=1}^m \lambda^{(j)}(\cdot)$ and for $x=0, \ldots, S-1$, $\Lambda \sum_{j=1}^m p^{(i)}(x+1)r^{(i)}(x+1) \leq \Lambda \sum_{j=1}^m p^{(i)}(x+1)r^{(i)}(x)\leq \Lambda \sum_{j=1}^m p^{(i)}(x)r^{(i)}(x)$ using the same arguments as in the inequalities of the previous point.
    Thus, $\nabla h^\pi(s+1)\leq \nabla h^\pi(s+2)$ implies that $\sum_{j=1}^m \lambda^{(j)}(s+1)\left[ r^{(j)}(s+1)-\nabla h^\pi(s+1) \right]^+ \geq \sum_{j=1}^m \lambda^{(j)}(s+2)\left[ r^{(j)}(s+2)-\nabla h^\pi(s+2) \right]^+$, which contradicts the initial inequality.
\end{itemize}
With the proofs of these three results modified as indicated, and state-dependent arrival rates, all results derive from the same original proofs.

\section{Value Iteration or Policy Iteration?}
\label{section: VI or PI}
\subsection{Policy Iteration for Admission Control}
We recall the definition of bias optimal policy:
\begin{definition}[\protect{bias optimal policy (see \citealp[definition 4.1]{feinbergOptimalityTrunkReservation2011})}]
    \label{def: bias optimal policy}
    A policy $\pi_b$ is bias optimal if it is gain optimal and its bias is maximal:
    \begin{equation*}
        \forall \pi \in \Pi, \forall s, h^\pi(s) \leq h^{\pi_b}(s)
    \end{equation*}
\end{definition}
A slight modification of the PI algorithm presented in Alg. \ref{alg: PI} can return a bias optimal policy (definition \ref{def: bias optimal policy}):

\begin{theorem} \label{theorem: pi}
    The PI algorithm converges to the bias optimal policy if the improvement step selects the policy which accepts the most job classes for each state.
    \label{thm: Policy Iteration for bias optimality}
\end{theorem}

To present the intuition behind this theorem, we first need to introduce trunk reservation policies.

\begin{definition}[from \protect{\citealp[definition 1.1]{feinbergOptimalityTrunkReservation2011}}]
    A policy $\pi$ is called a trunk reservation policy (TRP) if there are $m$ control levels $M^\pi_i, i=1 \ldots, m$, one per job class, such that a class $i$ arrival is admitted to the system if and only if there are less than $M^\pi_i$ jobs already in the system.
\end{definition}

Feinberg and Yang's theorem 4.1 states that the bias optimal policy $\pi^b$ exists, is unique and is the TRP with the largest optimal control levels for each job class. 
Furthermore, from their theorem 3.5, we know that this policy satisfies equation \eqref{eq: Bellman equation}.
Because PI returns policies satisfying \eqref{eq: Bellman equation} we propose in the improvement step to update the policy $\pi$ by accepting in state $s$ all job classes satisfying $r^{(i)}(s)\geq \nabla h^\pi(s)$. We show in the proof of Thm \eqref{theorem: pi} that at convergence, the policy obtained is the gain optimal one accepting the most job classes in each state.

\paragraph{Proof of theorem \ref{thm: Policy Iteration for bias optimality}}
Recall that the improvement step in PI consists in selecting better actions.
We only propose a rule to select the new policy among the possible improvements, which means that the algorithm still converges.
Let us call $\pi^c$ the policy obtained after convergence.
We want to prove that it is equal to the bias optimal policy $\pi^b$.
To lighten the notations, we note $h^c$ the bias of policy $\pi^c$ and $h^b$ the bias of policy $\pi^b$.
\begin{align*}
    i\in \pi^c(s) &\Leftrightarrow r^{(i)}(s) \geq \nabla h^c(s) && \text{because $\pi_c$ satisfies \eqref{eq: Bellman equation} and reciprocally because of the proposed improvement rule.}\\
    &\Leftrightarrow r^{(i)}(s) \geq \nabla h^b(s) && \text{since $\nabla h^c=\nabla h^b$ according to Feinberg and Yang's lemma 3.7 \citeyearpar{feinbergOptimalityTrunkReservation2011}} \\
    &\Leftrightarrow i\in \pi^b(s) &&\text{as $\pi^b$ is the optimal TRP with the largest control levels and reciprocally with \eqref{eq: Bellman equation}.}
\end{align*}
We precise the justification of the last equivalence: if $i\notin \pi^b(s)$, then either $\pi^b$ is not a TRP or $\pi^c$ is a TRP with larger control levels.

We know that variant of PI converges to the bias optimal policy. 
Any policy proposed by PI should satisfy that if $r^{(i)}(s) \geq r^{(j)} (s)$ in state $s$ and job class $j$ is accepted, then class $i$ must also be accepted. This restricts the number of potential optimal policies to $(m+1)^S$. 
However, to the best of our knowledge, the convergence speed of PI algorithms in this context is not known.

\subsection{Value Iteration for Admission Control}
Traditionally, UCRL2 uses Extended Value Iteration to identify the optimistic MDP and to derive a policy whose average reward matches the optimal average reward of the optimistic MDP, with a specified level of precision.
Since the optimistic CTMDP is known in our setting, we can replace the extended VI by classical VI.

As explained in section \ref{section: background}, VI is applied to an MDP to compute a sequence of bias vectors $(u)$. 
In our case, since we uniformized formulation of the problem, the stopping criterion becomes $span(u_{l+1}(s) - u_l) < \varepsilon/U$, where $U$ is the uniformization rate used to transform the CTMDP into an MDP. Indeed, when VI stops, we know that for the uniformized system, $\rho^{VI}_U \geq \rho^*_U - \varepsilon/U$ where $\rho^{VI}_U = \rho^{VI}/U$ is the average reward for the final policy of VI and $\rho^*_U = \rho^* / U$ is the optimal average reward. A simple multiplication by $U$ shows the desired inequality $\rho^{VI}\geq \rho^* - \varepsilon$.
When applied to the optimistic CTMDP $\tilde{\mathcal{M}}_k$ with precision $\varepsilon_k$, we obtain $\rho_k^{VI} \geq \tilde{\rho}_k - \varepsilon_k$, where $\tilde{\rho}_k$ represents the average reward of $\tilde{\mathcal{M}}_k$. 
Furthermore, under the assumption that the true CTMDP $\mathcal{M}$ lies in $\mathcal{C}_k$, $\rho_k^{VI}\geq \rho^* - \varepsilon_k$.

The convergence speed of VI is geometric, as stated in part \ref{section: background}, 
and it can be accelerated by a thoughtful initialization of the bias vector $u_0$. 
For example, in episode $k$, we propose to initialize it with the value function $h^{\pi_{k-1}}_k$ of the last policy computed for the new optimistic arrival rates.
This can be done by evaluating $\nabla h^{\pi_{k-1}}_k$ with equations \eqref{eq: closed-form expression for the average reward} and \eqref{eq: closed-form expression for the relative bias}. 
Then, we can choose $u_0(0)=0$ and for all $s>0$, $u_0(s)=-\sum_{x=0}^{s-1}\nabla h^{\pi_{k-1}}(x)$.
Thus, $u_0 = h_k^{\pi_{k-1}}-h_k^{\pi_{k-1}}(0)$
Intuitively, the policy updates given by VI after each episode should not be tolarge,, 
so the same should hold for the bias (up to an additive constant).
Hence, we hope that this initialization is not too far from a bias satisfying the stopping criterion. 

We adapt VI to the admission control problem in Alg. \ref{alg: VI}. 
Usually, improvement and evaluation are combined in a single step.
But we separate them to use the same improvement step as in PI. This eliminates the need to compute transition probability matrices for all possible actions.

\section{Accelerating Policy Iteration}

\label{appendix: E}
\begin{lemma}
    \label{lemma: closed-form expression for the relative bias}
    For $s \in \{ 0, 1, \dots, S-1 \}$, the relative bias $\nabla h^\pi(s)$ satisfies 
    \begin{equation}
        \label{eq:nabla h}
        \mu(s+1)\nabla h^\pi(s) = \sum_{q=s+1}^{S} \left(\rho^\pi - R^\pi(q) \right) \prod_{p=s+1}^{q-1} \frac{\Lambda^\pi(p)}{\mu(p+1)}
    \end{equation}
\end{lemma}

\label{proofs: closed-form formulas}
\paragraph{Proof of lemma \ref{lemma: closed-form expression for the relative bias}}
We will prove equation~\eqref{eq:nabla h} by induction on the state $s$.

\textit{Initialization:} equation \eqref{eq: average reward for admission control} for $s=S$ yields $\rho^\pi=\mu(S)\nabla h^\pi(S-1)$, since the system is full and all incoming jobs are refused.
Formula~\eqref{eq:nabla h} is therefore true for $s=S-1$.

\textit{Induction step:} suppose formula \eqref{eq:nabla h} is true for $s+1$, with $0 \leq s \leq S-2$. We want to prove that it is true for $s$ as well.
Equation~\eqref{eq: average reward for admission control} for $s+1$ gives
\begin{equation}
    \rho^\pi = R^\pi(s+1) - \Lambda^\pi(s+1) \nabla h^\pi(s+1) + \mu(s+1)\nabla h^\pi(s) \label{eq: step0}
\end{equation}
where we remind that $R^\pi(s)=\sum_{i \in \pi(s)}\lambda^{(i)} r^{(i)}(s)$ and $\Lambda^\pi(s)=\sum_{i \in \pi(s)}\lambda^{(i)}$. Since $s>0$$, \mu(s+2)\neq 0$, so we obtain \eqref{eq: step1} by multiplying $\Lambda^\pi(s+1)\nabla h^\pi(s+1)$ in \eqref{eq: step0} by $\mu(s+2)/\mu(s+2)$ . We then inject formula~\eqref{eq:nabla h}, true for $s+1$, to obtain \eqref{eq: step2}. Moving $\Lambda^\pi(s+1)/\mu(s+2)$ in the product operator gives \eqref{eq: step3}.
\begin{align}
    \rho^\pi &= R^\pi(s+1) - \Lambda^\pi(s+1) \nabla h^\pi(s+1)\frac{\mu(s+2)}{\mu(s+2)} + \mu(s+1)\nabla h^\pi(s) \label{eq: step1}\\
    &= R^\pi(s+1) - \frac{\Lambda^\pi(s+1)}{\mu(s+2)} \sum_{q=s+2}^S(\rho^\pi-R^\pi(q))\prod_{p=s+2}^{q-1}\frac{\Lambda^\pi(p)}{\mu(p+1)} + \mu(s+1)\nabla h^\pi(s) \label{eq: step2}\\
    &= R^\pi(s+1) - \sum_{q=s+2}^S(\rho^\pi-R^\pi(q))\prod_{p=s+1}^{q-1}\frac{\Lambda^\pi(p)}{\mu(p+1)} + \mu(s+1)\nabla h^\pi(s) \label{eq: step3}
\end{align}
By isolating $\mu(s+1)\nabla h^\pi(s)$ and rearranging the terms, we finally get:
\begin{align*}
    \mu(s+1)\nabla h^\pi (s) &= \rho^\pi - R^\pi(s+1)  + \sum_{q=s+2}^{S} \left( \rho^\pi-R^\pi(q) \right) \prod_{p=s+1}^{q-1}\frac{\Lambda^\pi(p)}{\mu(p+1)} \\
    &= (\rho^\pi - R^\pi(s+1)) \prod_{p=s+1}^s \frac{\Lambda^\pi(p)}{\mu(p+1)} + \sum_{q=s+2}^{S} \left( \rho^\pi-R^\pi(q) \right) \prod_{p=s+1}^{q-1}\frac{\Lambda^\pi(p)}{\mu(p+1)} 
\end{align*}
which proves, after moving the terms on the right-hand side under the same sum, that the formula~\eqref{eq:nabla h} is also true for $s$.

\paragraph{Proof of proposition \ref{prop: closed-form expression as a matrix product}}
    We note $\nabla H^\pi$ the column vector of size $S$ such that ${\left(\nabla H^\pi\right)}_s = \nabla h^\pi(s) = h^\pi(s)-h^\pi(s+1)$, $R^\pi$ the vector of reward per second of size $S$ with
${\left(R^\pi\right)}_q=R^\pi(q)$, $\bm{e}$ the column vector composed of $S$ ones
and $U^\pi$ is the square matrix of size $S$ such that 
${\left(U\right)}_{s,q} = \mathds{1}_{\left\{ q\geq s+1 \right\}}\frac{1}{\mu(s+1)}\prod_{p=s+1}^{q-1} \frac{\Lambda^\pi(p)}{\mu(p+1)}$. Let us show that $\nabla H^\pi = U^\pi(\rho^\pi \bm{e}-R^\pi)$.
\begin{align*}
    \left(U^\pi\left(\rho^\pi \bm{e}-R^\pi\right)\right)(s) &= \frac{1}{\mu(s+1)}\sum_{q=s+1}^{S} (\rho^\pi-R^\pi(q))\prod_{p=s+1}^{q-1} \frac{\Lambda^\pi(p)}{\mu(p+1)} \\
    &= \nabla h^\pi(s) \quad \text{according to lemma \ref{lemma: closed-form expression for the relative bias}} 
\end{align*}

\section{Regret Analysis}
\label{appendix: regret}

For lemmas $\ref{lemma: Rout}$, $\ref{lemma: Rbad}$, $\ref{lemma: Rgood}$ and theorem $\ref{thm: upper bound of the expected regret}$, we suppose that the immediate rewards $R_1, \ldots, R_m$ are bounded by $R_{\max}$.
The constant total arrival rate $\Lambda$ allows to separate the influence of the number of classes from that of the system load.

\subsection{Lemmas}
In the following two lemmas, we provide bounds for terms $\mathcal{R}_{(K)}^{out}$ and $\mathcal{R}_{(K)}^{bad}$.

\begin{lemma}
    \label{lemma: Rout}
    With $\delta_k=1/(\mu t_k)$,
    the total expected regret for all episodes $k=1, \dots, K$ in which the true CTMDP $\mathcal{M}$
    does not belong to the confidence set $\mathcal{C}_k$
    is upper bounded:
    \begin{equation*}
        \mathcal{R}_{(K)}^{out} \leq 4\frac{K \rho^*}{\mu}+ \rho^* t_1
    \end{equation*}
\end{lemma}

\begin{lemma}
    \label{lemma: Rbad}
    The total expected regret from episodes $k=1, \dots, K$ in which there is a job class such that $N^{(i)}_k$ is smaller than $t_k\lambda^{(i)}/2$
    is upper bounded:
    \begin{equation*}
        \mathcal{R}_{(K)}^{bad} \leq  14 \frac{K \rho^* }{\Lambda}
    \end{equation*}
\end{lemma}

We then have to bound the main term $\mathcal{R}_{(K)}^{good} = \E{\sum_{k=1}^K E_k}$ where $E_k=\E{\left.\Delta_k^{good}\middle| t_k(\cdot) \right.}$ is the expected regret conditioned on the total time $t_k(s)$ spent in each state $s$ in episode $k$. 
When the total sojourn times $t_k(s), s=0 \ldots, S$ are known,
the number $\nu_k(s,i)$ of accepted jobs of class $i$ in state $s$ for episode $k$ follows a Poisson distribution of parameter $\lambda^{(i)} t_k(s)$, under the condition that class $i$ is accepted in state $s$. Otherwise, it is null.
More precisely, we have
\begin{align*}
    E_k &= \E{\left.\Delta_k^{good}  \middle|  t_k(\cdot)\right.} = \E{\left. t_k \rho^* - \sum_{s,i \in \pi_k(s)}\nu_k(s,i) r^{(i)}(s) \middle| t_k(\cdot) \right.} \\
    &= \sum_s t_k(s) \left( \rho^* - \sum_{i \in \pi_k(s)}\lambda^{(i)} r^{(i)}(s)\right) &&\text{as }\nu_k(s,i) \sim \mathcal{P}(\lambda^{(i)} t_k(s)) \text{ when} \\
    & && t_k(s) \text{ is known.} \\
    &= \underbrace{\sum_s t_k(s) (\rho^* - \tilde{\rho}_k)}_{(A)} + \underbrace{\sum_s t_k(s) \left( \tilde{\rho}_k - \tilde{R}_k(s) \right)}_{(B)} + \underbrace{\sum_s t_k(s) \left(\tilde{R}_k(s) - R_k(s) \right)}_{(C)} && \tilde{R}_k(s) = \sum_{i\in \pi_k(s)} \tilde{\Lambda}_k \tilde{p}_k^{(i)}(s) r^{(i)}(s) \\
    & && \text{ and }R_k(s) = \sum_{i\in \pi_k(s)} \lambda^{(i)} r^{(i)}(s)  
\end{align*}

In the following lemma, we give an upper bound to $\mathcal{R}_{(K)}^{good}$ by bounding $(A), (B)$ and $(C)$. 
We deal with both PI and VI.
Indeed, only the bound of $\mathcal{R}_{(K)}^{good}$ is impacted by the choice of PI or VI, whereas the bounds of $\mathcal{R}_{(K)}^{out}$ and $\mathcal{R}_{(K)}^{bad}$ do not depend on it.

\begin{lemma}
    \label{lemma: Rgood}
    With Policy Iteration,
    \begin{equation*}
        \begin{aligned}
        \mathcal{R}_{(K)}^{good} &\leq KS\frac{\Lambda_{\max}R_{\max}}{\mu_{\max}} + 4\sqrt{T_K \log (2\mu T_K)} \left(4\frac{\Lambda_{\max}^2}{\Lambda_{\min}\sqrt{\Lambda}}+ \sqrt{m\Lambda}\right)\left(1+\frac{\Lambda_{\max}}{\mu_{\max}}\right)R_{\max}
    \end{aligned}
    \end{equation*}

    With Value Iteration,
    \begin{equation*}
        \begin{aligned}
        \mathcal{R}_{(K)}^{good} &\leq KSV + KR_{\max} + 4\sqrt{T_K \log (2\mu T_K)} \left(4\frac{\Lambda_{\max}^2}{\Lambda_{\min}\sqrt{\Lambda}}+ \sqrt{m\Lambda}\right)\left(R_{\max}+V\right)
    \end{aligned}
    \end{equation*}
    with 
    \begin{equation}
        \label{eq: expression of V}
        V = \underset{s}{\max}\left\{\frac{\Lambda_{\max} R_{\max}}{\mu} \sum_{q=s+1}^S \prod_{p=s+1}^{q-1} \frac{\tilde{\Lambda}^\pi(p)}{\mu(p+1)}\right\}
    \end{equation}
\end{lemma}
\subsection{Proofs}
\paragraph{Proof of Lemma \ref{lemma: Rout}}
To bound $\mathcal{R}_{(K)}^{out} = \E{\sum_{k=1}^K\Delta_k \mathds{1}_{\mathcal{M}\notin\mathcal{C}_k}}$, we need to compute the probability for the true CTMDP $\mathcal{M}$ to be outside of the confidence set $\mathcal{C}_k$ after the update phase in episode $k$. 
We recall that a CTMDP $M$ belongs to $\mathcal{C}_k$ if and only if $\Lambda\in CI_\Lambda(k)$ and $p\in CI_p(k)$.

\begin{align*}
    \mathbb{P}(\mathcal{M}\notin \mathcal{C}_k) &= \mathbb{P}(\Lambda \notin CI_\Lambda(k) \cup p \notin CI_p(k)) && \text{by definition of $\mathcal{C}_k$}\\ 
    &\leq \mathbb{P}\left(\Lambda \notin CI_\Lambda(k)\right) + \mathbb{P}\left(p \notin CI_p(k)\right)  \\
    &= \mathbb{P}\left(\Lambda \notin \left[\Lambda_{\min}, \Lambda_{\max}\right]\cap \left[ \hat{\Lambda}_k - \varepsilon_k, \hat{\Lambda}_k + \varepsilon_k \right]\right) + \mathbb{P}\left(p \notin CI_p(k)\right) && \text{by definition of $CI_\Lambda(k)$} \\
    &= \mathbb{P}\left(\Lambda \notin  \left[ \hat{\Lambda}_k - \varepsilon_k, \hat{\Lambda}_k + \varepsilon_k \right]\right) + \mathbb{P}\left(p \notin CI_p(k)\right) && \text{by assumption} \\
    &\leq 2\delta_{k-1} && \text{with lemma \ref{proposition: confidence intervals}}
\end{align*}

We can now bound $\mathcal{R}_{(K)}^{out}$. Recall that the regret of episode $k$ is $\Delta_k = t_k \rho^* - \sum_{s,i}\nu_k(s,i)r^{(i)}(s)$,
where $\nu_k(s,i)$ is the number of job of class $i$ that are accepted while the system is in state $s$ during episode $k$. 
Since the distribution $\tilde{p}_1$ can be arbitrarily far from the true one, 
we take care of the regret of the first episode in $\mathcal{R}_1^{out}$.
Assuming that no job of class $i$ is accepted in state $s$ if $r^{(i)}(s)<0$,
we derive the following upper bound:
\begin{align*}
    \mathcal{R}_{(K)}^{out} &= \E{\sum_{k=1}^K \Delta_k \mathds{1}_{\mathcal{M}\notin \mathcal{C}_k}} && \text{by definition}\\
    &\leq \rho^* t_1+\E{\sum_{k=2}^K  \mathds{1}_{\mathcal{M}\notin \mathcal{C}_k}  \rho^* t_k} && \text{by removing the rewards, which are non-negative}\\
    &\leq \rho^* t_1 + \rho^* \sum_{k=2}^K t_k \mathbb{P}\left( \mathcal{M}\notin \mathcal{C}_k \right) && \text{by taking the deterministic terms out of the expectation}\\
    &\leq \rho^* t_1 + 2\rho^* \sum_{k=2}^K t_k  \delta_{k-1} && \text{from the previous computations}
\end{align*}

With $\delta_{k-1}=\frac{1}{\mu t_{k-1}}=\frac{2}{\mu t_{k}}$, we obtain $\mathcal{R}_{(K)}^{out} \leq  4\frac{K \rho^*}{\mu}+ \rho^* t_1$.

\paragraph{Proof of Lemma \ref{lemma: Rbad}}
We note $\nu_k$ the number of arrivals in episode $k$.
Following the same steps as in the proof of lemma \ref{lemma: Rout}, we have
\begin{equation*}
     \mathcal{R}_{(K)}^{bad} \leq \rho^* \sum_{k=2}^K t_k \mathbb{P}\left(\nu_{k-1} < \frac{\Lambda t_{k-1}}{2}\right)
\end{equation*}

 Episode $k$ is of length $t_k$ and the job arrivals follow a Poisson process of rate $\Lambda$, so $\nu_k$ is a Poisson random variable with parameter $\Lambda t_k$. The probability of $\nu_k$ being smaller than $\Lambda t_k/2$ is
\begin{equation*}
    \mathbb{P}\left( \nu_k < \Lambda t_k/2 \right) = \sum_{p=0}^{\lceil \Lambda t_k/2 \rceil-1} e^{-\Lambda t_k} \frac{(\Lambda t_k)^p}{p!}
\end{equation*}
We compute an upper bound of this probability:
\begin{align*}
    \sum_{p=0}^{\lceil \Lambda t_k/2 \rceil-1} e^{-\Lambda t_k} \frac{(\Lambda t_k)^p}{p!} & =  \sum_{p=0}^{\lceil \Lambda t_k/2 \rceil-1} e^{-\Lambda t_k} \frac{(\Lambda t_k/2)^p}{p!} 2^p && \text{multiplying by } \frac{2^p}{2^p} \\
    & \leq  \sum_{p=0}^{\lceil \Lambda t_k/2 \rceil-1} e^{-\Lambda t_k} \frac{(\Lambda t_k/2)^p}{p!} 2^{\Lambda t_k/2} && \text{since }p < \Lambda t_k/2 \\
    & \leq  2^{\Lambda t_k/2} e^{-\Lambda t_k} \sum_{p=0}^{\infty}  \frac{(\Lambda t_k/2)^p}{p!} && \text{extending the sum to all $p$}  \\
    & \leq 2^{\Lambda t_k/2}e^{-\Lambda t_k/2} && \text{rewriting the sum as an exponential}\\
    & = \exp\left({-\frac{1-\log 2}{2}\Lambda t_k }\right) &&\text{rewriting as a product of exponentials}
\end{align*}

We can now bound $\mathcal{R}_{(K)}^{bad}$:
\begin{align*}
    \mathcal{R}_{(K)}^{bad} & \leq \rho^* \sum_{k=2}^K t_k \mathbb{P}\left(\nu_{k-1} < \frac{\Lambda t_{k-1}}{2}\right)                         \\
    & \leq \rho^* \sum_{k=2}^K t_k \exp\left( -\frac{1-\log 2}{2}\Lambda t_{k-1} \right) && \text{injecting the previous result} \\
    & \leq \rho^* \sum_{k=2}^K \frac{t_k}{\frac{1-\log 2}{2}\Lambda t_{k-1}}  && \text{as } ue^{-u} \leq 1 \text{ for } u \geq 0  \\
    & \leq \rho^* \sum_{k=2}^K \frac{2}{\frac{1-\log 2}{2}\Lambda}  && \text{since } t_k = 2t_{k-2}  \\
    & < 14 \frac{K \rho^* }{\Lambda} && \text{summing over $k$ and simplifying with $4/(1-\log 2) < 14$}
\end{align*}

\paragraph{Proof of Lemma \ref{lemma: Rgood}}
We assume that the number arrivals $\nu_k$ in episode $k$ satisfies $\nu_k \geq \Lambda t_k / 2$. We also suppose that the true CTMDP $\mathcal{M}\in \mathcal{C}_k$, which implies that the true optimal average reward $\rho^*$ is smaller than the optimal average reward $\tilde{\rho}_k$ of the optimistic CTMDP in $\mathcal{C}_k$: $\tilde{\rho}_k \geq \rho^*$. We note $\nu_k(s,i)$ the number of class $i$ jobs that are accepted in state $s$ by following policy $\pi_k$. We want to bound the following terms:
\begin{align*}
    (A) &= \sum_s t_k(s) \left(\rho^* - \tilde{\rho}_k \right) \\
    (B) &= \sum_s t_k(s) \left( \tilde{\rho}_k - \tilde{R}_k(s) \right)\\
    (C) &= \sum_s t_k(s) \left(\tilde{R}_k(s) - R_k(s) \right)
\end{align*}

We bound them for PI. We will explain afterwards what differs with VI.

\textit{Bounding $(A)$.}
Since $\tilde{\rho}_k$ is the optimal average reward of the optimistic CTMDP in $\mathcal{C}_k$ and the true CTMDP $\mathcal{M}$ belongs to $\mathcal{C}_k$, we know that $\rho^* \leq \tilde{\rho}_k$ so that $(A) \leq 0$.

\textit{Bounding $(B)$.}
Recall from equation \eqref{eq: fixed-point equation} that $\tilde{\rho}_k$ and $\tilde{h}_k$ satisfy $\tilde{\rho}_k-\tilde{R}_k = \tilde{Z_k} \tilde{h}_k$, so that we can decompose $(B)$ in two terms:
\begin{equation*}
    (B)=\sum_s t_k(s) \left( \tilde{\rho}_k - \tilde{R}_k(s) \right) = \bm{t_k}^T \tilde{Z}_k \tilde{h}_k 
    = \bm{t_k}^T \left(\tilde{Z}_k - Z_k\right) \tilde{h}_k + \bm{t_k}^T Z_k \tilde{h}_k
\end{equation*}
where $\bm{t_k}$ is the vector of composed of $t_k(s), s=0, \dots, S$, $Z_k$ is the infinitesimal generator of the true CTMDP and $\tilde{Z}_k$ is the infinitesimal generator of the optimistic CTMDP in $\mathcal{C}_k$, both following the same policy $\pi_k$.

We will bound these two terms separately. Let us begin with 
\begin{align*}
    \bm{t_k}^T \left(\tilde{Z}_k - Z_k\right) \tilde{h}_k &= \sum_s t_k(s) \sum_{i \in \pi_k(s)}\left( \tilde{\lambda}_k^{(i)}(s) - \lambda^{(i)} \right) \nabla \tilde{h}_k(s)\\
    &= \sum_s t_k(s) \sum_{i \in \pi_k(s)}\left( \tilde{\Lambda}_k\tilde{p}_{k}^{(i)}(s) - \Lambda p^{(i)} \right) \nabla \tilde{h}_k(s)  \\
    &=\sum_s t_k(s) \sum_{i \in \pi_k(s)}\left(\left(\tilde{\Lambda}_k-\Lambda\right) \tilde{p}^{(i)}_k(s) + \Lambda \left( \tilde{p}_{k}^{(i)}(s)-p^{(i)}\right) \right) \nabla \tilde{h}_k(s)  \\
    &\leq \sum_s t_k(s) \left( \tilde{\Lambda}_k-\Lambda  + \Lambda\lVert \tilde{p}^{(k)}(s) - p \rVert_1\right)\nabla\tilde{h}_k(s)
\end{align*}

% We apply Theorem \ref{theorem: bound for relative bias} to precise the bound:

With PI, we know that we obtain an optimal policy. We can apply Theorem \ref{theorem: bound for relative bias}:

\begin{align*}
    \bm{t_k}^T \left(\tilde{Z}_k - Z_k\right) \tilde{h}_k &\leq t_k \left(\tilde{\Lambda}_k-\Lambda  + \Lambda\lVert \tilde{p}^{(k)} - p \rVert_1\right)\frac{\Lambda_{\max}R_{\max}}{\mu_{\max}} && \text{by bounding $\nabla\tilde{h}_k(s)$ and summing over $s$}\\
    &\leq 2t_k\left( \varepsilon_{k,\Lambda}+\Lambda\varepsilon_{k,p} \right)\frac{\Lambda_{\max}R_{\max}}{\mu_{\max}} && \text{$\tilde{\Lambda}_k - 2\varepsilon_{k,\Lambda} \leq \Lambda \leq \tilde{\Lambda}_k$, same idea for $p$} 
\end{align*}
\begin{align*}
\bm{t_k}^T \left(\tilde{Z}_k - Z_k\right) \tilde{h}_k &
\leq 2t_k \left( 4\frac{\Lambda_{\max}^2}{\Lambda_{\min}}\sqrt{\frac{2}{\nu_{k-1}}\log\frac{1}{\delta_{k-1}}} + \Lambda \sqrt{\frac{2m}{N_{k-1}}\log \frac{2}{\delta_{k-1}}} \right)\frac{\Lambda_{\max}R_{\max}}{\mu_{\max}} \\
    &\leq 4\sqrt{2}\sqrt{t_k} \left( 4\frac{\Lambda_{\max}^2}{\Lambda_{\min}}\sqrt{\frac{1}{\Lambda}\log\frac{1}{\delta_{k-1}}} + \sqrt{m\Lambda\log \frac{2}{\delta_{k-1}}} \right)\frac{\Lambda_{\max}R_{\max}}{\mu_{\max}} 
\end{align*}
since $N_{k-1} \geq \nu_{k-1} \geq \Lambda t_{k-1}/2$.

We now sum over $k=2$ to $K$, since the case $k=1$ is covered by $\mathcal{R}_1^{out}$. First, remark that we can bound $\sum_{k=2}^K\sqrt{t_k}$:
\begin{equation*}
    \sum_{k=2}^K\sqrt{t_k} =  \sum_{k=2}^K\sqrt{t_1 2^{k-2}} = \left(\sqrt{2} + 1\right) \left(\sqrt{T_k} - \sqrt{t_1} \right) \leq \left(\sqrt{2} + 1\right) \sqrt{T_k}
\end{equation*}

Thus, 
\begin{equation}
    \label{eq: PI_B lemma 3 res 1}
    \sum_{k=2}^K \bm{t_k}^T \left(\tilde{Z}_k - Z_k\right) \tilde{h}_k \leq 
    14\sqrt{T_K} \left( 4\frac{\Lambda_{\max}^2}{\Lambda_{\min}}\sqrt{\frac{1}{\Lambda}\log \mu T_K} + \sqrt{m\Lambda\log 2\mu T_K} \right)\frac{\Lambda_{\max}R_{\max}}{\mu_{\max}}
\end{equation}

We now have to bound $\bm{t_k}^T Z_k \tilde{h}_k$. 
We want to express it as a sum over a trajectory.
First, we rewrite this equation for the uniformized system:
\begin{equation*}
    \bm{t_k}^TZ_k \tilde{h}_k = U\bm{t_k}^T \left(P_k - I\right) \tilde{h}_k
\end{equation*}
where $U$ is a uniformization constant and $P_k$ is the corresponding transition probability matrix: 
\begin{equation*}
    P_k = I + \frac{1}{U} Z_k
\end{equation*}
$Ut_k(s)$ is the number of times the uniformized MDP visits state $s$ and $Ut_k$ is the total number of visits in episode $k$. Remark that since we use the uniformization as a computation trick only, we can choose a constant $U$ such that $Ut_k(s)\in \mathbb{N}$ for all state $s$. 
We can now decompose that sum over a trajectory $(s_1, \ldots, s_{Ut_k})$ like Anselmi et al. in appendix A.3.5 \citeyearpar{anselmiReinforcementLearningBirth2022}:
\begin{align*}
    U\bm{t_k}^T \left(P_k - I\right) \tilde{h}_k &= \sum_{\tau=1}^{Ut_k} \left(P_k(\cdot | s) - \bm{e}_s \right)\tilde{h}_k \\
    &= \tilde{h}_k(s_{Ut_k+1}) - \tilde{h}_k(s_1) + \sum_{\tau=1}^{Ut_k} \Phi_\tau
\end{align*}
with $\bm{e}_s$ the column vector with one in position $s$ and zeros everywhere else, $\Phi_\tau = \left(P_k(\cdot | s_\tau) - e_{s_{\tau+1}}\right) \tilde{h}_k$.
$\Phi_\tau = \Phi_1, \Phi_2, \ldots, \Phi_{\nu_k}$ is a martingal difference as $\E{\Phi_\tau|s_1, \ldots, s_\tau}=0$. Indeed, 
\begin{align*}
    \E{\Phi_\tau|s_1, \ldots, s_\tau} &= P_k(\cdot | s_\tau) \tilde{h}_k - \E{e_{s_{\tau+1}}| s_1, \dots, s_\tau}\tilde{h}_k \\
    &= P_k(\cdot | s_\tau) \tilde{h}_k - P_k(\cdot | s_\tau) \tilde{h}_k
\end{align*}
so that the expectation of $\sum_{\tau=1}^{U t_k}\Phi_\tau$ is equal to zero.

For PI, $\pi_k$ is gain optimal for the optimistic CTMDP $\tilde{M}_k \in \mathcal{C}_k$. Lemma \ref{theorem: bound for relative bias} tells us that $\tilde{h}_k$ is a decreasing function of $s$. Then,
\begin{itemize}
    \item if  $s_{\nu_k+1} \geq s_{1}$,  $\tilde{h}_k(s_{U t_k+1}) - \tilde{h}_k(s_1)\leq 0$;
    \item otherwise, $\tilde{h}_k(s_{U t_k+1}) - \tilde{h}_k(s_1) = \sum_{x=s_{Ut_k+1}}^{s_1} \nabla\tilde{h}_k(x)$
\end{itemize}
so that using lemma \ref{theorem: bound for relative bias} again,
\begin{equation*}
    \tilde{h}_k(s_{U t_k+1}) - \tilde{h}_k(s_1) \leq S\frac{\Lambda_{\max} R_{\max}}{\mu_{\max}}
\end{equation*}

Finally, summing over $k$ yields 
\begin{equation}
    \label{eq: PI_B lemma 3 res 2}
    \sum_{k=2}^K \bm{t_k}^T Z_k \tilde{h}_k \leq KS\frac{\Lambda_{\max} R_{\max}}{\mu_{\max}}
\end{equation}

\textit{Bounding $(C)$.}
The following development is true for both PI and VI. The only difference is that the policies may differ.

\begin{align*}
     \sum_s t_k(s) \left(\tilde{R}_k(s) - R_k(s) \right) &= \sum_s t_k(s) \sum_{i\in \pi_k(s)} \left( \tilde{\lambda}^{(i)}_k - \lambda^{(i)}\right)r^{(i)}(s)\\
     &= \sum_s t_k(s) \sum_{i\in \pi_k(s)} \left[ \left(\tilde{\Lambda}-\Lambda\right)\tilde{p}^{(i)} + \Lambda \left(\tilde{p}^{(i)}-p^{(i)}\right) \right]r^{(i)}(s) \\
     &\leq t_k \left( \tilde{\Lambda}-\Lambda + \Lambda \lVert\tilde{p}^{(k)} - p\rVert \right)R_{\max} \\
     &\leq 2t_k \left( \varepsilon_{k,\Lambda} + \Lambda \varepsilon_{k, p} \right) R_{\max} \\
     &\leq 2 t_k \left( 4\frac{\Lambda_{\max}^2}{\Lambda_{\min}}\sqrt{\frac{2}{\nu_{k-1}}\log\frac{1}{\delta_{k-1}}} + \Lambda \sqrt{\frac{2m}{N_{k-1}}\log \frac{2}{\delta_{k-1}}} \right) R_{\max}\\
     &\leq 4\sqrt{2}\sqrt{t_k}\left( 4\frac{\Lambda_{\max}^2}{\Lambda_{\min}}\sqrt{\frac{1}{\Lambda} \log \mu t_k} + \sqrt{m\Lambda \log 2 \mu t_k} \right)R_{\max} 
\end{align*}
so that by summing over $k=2$ to $K$:
\begin{equation}
    \label{eq: PI_B lemma 3 res 3}
   \sum_{k=2}^K \sum_s t_k(s) \left( \tilde{R}_k(s) - R_k(s) \right) \leq 14\sqrt{T_K}\left( 4\frac{\Lambda_{\max}^2}{\Lambda_{\min}}\sqrt{\frac{1}{\Lambda} \log \mu T_K} + \sqrt{m\Lambda \log 2 \mu T_K} \right)R_{\max} 
\end{equation}

We now describe what differs with VI:
\begin{itemize}
    \item for (A), we note $\tilde{\rho}^{VI}_k$ the average reward for the optimistic CTMDP and the policy $\pi^{VI}_k$ obtained by VI run with precision $\varepsilon_k$. 
We know that $\tilde{\rho}^{VI}_k \geq \tilde{\rho}_k-\varepsilon_k \geq \rho^* - \varepsilon_k$, again with the assumption that for episode $k$,  the true CTMDP $\mathcal{M}$ is in $\mathcal{C}_k$, so that $(1)\leq t_k \tilde{\rho}_k^{VI} + t_k \varepsilon_k$.
Choosing $\varepsilon_k = R_{\max}/t_k$, the term $t_k\varepsilon_k$ adds to the bound when summing over all the episodes: 
\begin{equation}
    \label{eq: VI_B lemma 3 res 1}
    K R_{\max}
\end{equation}

\item for (B), with VI, point $(2)$ of lemma \ref{theorem: bound for relative bias} cannot be applied as we have no guarantee that the policy is gain optimal when stopping.
However, its first point results in $\tilde{\rho}^{VI}_k \leq \Lambda_{\max}R_{\max}$, as $\tilde{\rho}^{VI}_k \leq \tilde{\rho}_k$.
From \ref{lemma: closed-form expression for the relative bias} applied to $\tilde{\mathcal{M}}_k$, we obtain the following bound by removing the reward per second, which is non-negative:
\begin{equation*}
    \left|\nabla \tilde{h}^{VI}_k(s) \right|
    \leq \frac{\tilde{\rho}^{VI}_k}{\mu} \sum_{q=s+1}^S \prod_{p=s+1}^{q-1} \frac{\tilde{\Lambda}^\pi(p)}{\mu(p+1)}
    \leq \frac{\Lambda_{\max} R_{\max}}{\mu} \sum_{q=s+1}^S \prod_{p=s+1}^{q-1} \frac{\tilde{\Lambda}^\pi(p)}{\mu(p+1)}
\end{equation*}
To work with compact equations, we define a constant $V$ such that 
\begin{equation*}
    V = \underset{s}{\max}\left\{\frac{\Lambda_{\max}  R_{\max}}{\mu} \sum_{q=s+1}^S \prod_{p=s+1}^{q-1} \frac{\tilde{\Lambda}^\pi(p)}{\mu(p+1)}\right\}
\end{equation*}  

We obtain the following bound:
\begin{equation}
    \label{eq: VI_B lemma 3 res 3}
    \sum_{k=2}^K \bm{t_k}^T \left(\tilde{Z}_k - Z_k\right) \tilde{h}_k \leq
    14\sqrt{T_K \log(2\mu T_K)}\left( 4\frac{\Lambda_{\max}^2}{\Lambda_{\min}\sqrt{\Lambda}}+\sqrt{m\Lambda} \right)V
\end{equation}
Then, using $\left| \nabla \tilde{h}^{VI}_k \right|\leq V$, we have 
\begin{equation*}
    \tilde{h}^{VI}_k(s_{U t_k+1}) - \tilde{h}^{VI}_k(s_1) \leq SV
\end{equation*}
and summing over the episodes, we obtain 
\begin{equation}
    \label{eq: VI_B lemma 3 res 2}
    \sum_{k=2}^K \bm{t_k}^T Z_k \tilde{h}^{VI}_k \leq KSV
\end{equation}
Note that for the worst case $\mu < \Lambda_{\max}$, V can be upper bounded by
\begin{align*}
    V &\leq \frac{\Lambda_{\max}  R_{\max}}{\mu} \sum_{q=1}^S \prod_{p=1}^{q-1} \frac{\Lambda_{\max}}{\mu} \\
    &\leq \frac{\Lambda_{\max} R_{\max}}{\mu}\frac{\left(\frac{\Lambda_{\max}}{\mu}\right)^S-1}{\frac{\Lambda_{\max}}{\mu}-1}
\end{align*}
so that in the worst case, $V$ is upper bounded by a term that is exponential in $S$.
\end{itemize}

\textit{Conclusion.}\
 
With PI, it follows from \eqref{eq: PI_B lemma 3 res 1}, \eqref{eq: PI_B lemma 3 res 2} and \eqref{eq: PI_B lemma 3 res 3} that 
\begin{equation*}
    \begin{aligned}
        \mathcal{R}_{(K)}^{good} &\leq KS\frac{\Lambda_{\max}R_{\max}}{\mu_{\max}} + 14\sqrt{T_K \log (2\mu T_K)} \left(4\frac{\Lambda_{\max}^2}{\Lambda_{\min}\sqrt{\Lambda}}+ \sqrt{m\Lambda}\right)\left(1+\frac{\Lambda_{\max}}{\mu_{\max}}\right)R_{\max}
    \end{aligned}
\end{equation*}

With VI, it follows from \eqref{eq: PI_B lemma 3 res 3} (the bound still holds for VI), \eqref{eq: VI_B lemma 3 res 1}, \eqref{eq: VI_B lemma 3 res 3} and \eqref{eq: VI_B lemma 3 res 2} that 
\begin{equation*}
    \begin{aligned}
        \mathcal{R}_{(K)}^{good} &\leq KSV + KR_{\max} + 14\sqrt{T_K \log (2\mu T_K)} \left(4\frac{\Lambda_{\max}^2}{\Lambda_{\min}\sqrt{\Lambda}}+ \sqrt{m\Lambda}\right)\left(R_{\max}+V\right)
    \end{aligned}
\end{equation*}

\paragraph{Proof of Theorem \ref{thm: upper bound of the expected regret}}
By combining the results of the previous lemmas \ref{lemma: Rout}, \ref{lemma: Rbad} and \ref{lemma: Rgood}, the regret is upper bounded for PI by:
\begin{equation}
    \label{eq: final upper bound for PI}
    \begin{aligned}
        \mathcal{R}_{(K)} &\leq  \rho^* \left( t_1 + 4\frac{K}{\mu}+14\frac{K}{\Lambda} \right) + KS\frac{\Lambda_{\max}R_{\max}}{\mu_{\max}} + 14\sqrt{T_K \log (2\mu T_K)} \left(4\frac{\Lambda_{\max}^2}{\Lambda_{\min}\sqrt{\Lambda}}+ \sqrt{m\Lambda}\right)\left(1+\frac{\Lambda_{\max}}{\mu_{\max}}\right)R_{\max} 
    \end{aligned}
\end{equation}
and for VI by:
\begin{equation}
    \begin{aligned}
        \mathcal{R}_{(K)} &\leq \rho^*\left(t_1 + 4 \frac{K}{\mu} + 14\frac{K}{\Lambda}\right)+ KSV + KR_{\max} + 14\sqrt{T_K \log (2\mu T_K)} \left(4\frac{\Lambda_{\max}^2}{\Lambda_{\min}\sqrt{\Lambda}}+ \sqrt{m\Lambda}\right)\left(R_{\max}+V\right)
    \end{aligned}
\end{equation}

With $K=1+\log_2 \frac{T_K}{t_1}$,
we obtain for PI:
\begin{equation*}
    \begin{aligned}
        \mathcal{R}_{(K)} &\leq O\left( S\log T_K + \sqrt{mT_K\log T_K} \right) &&\text{for a finite number of servers}\\
        \mathcal{R}_{(K)} &\leq O\left( \log T_K + \sqrt{mT_K\log T_K} \right) &&\text{for an infinite number of servers, as $\mu_{\max}=S\mu$}
    \end{aligned}
\end{equation*}

For VI, we obtain:
\begin{equation*}
    \begin{aligned}
        \mathcal{R}_{(K)} &\leq  O\left( SV\log T_K + V\sqrt{mT_K\log T_K} \right)  && \text{for a finite number of servers}
    \end{aligned}
\end{equation*}

\section{Improving the Empirical Regret}
\label{appendix: empirical regret}
\paragraph{Proof of Proposition \ref{prop: learning lambda_max and lambda_min}}
We apply Lemma 1 of Bubeck et al. \citeyearpar{bubeckBanditsHeavyTail2013}:
for $k\geq 2$, with probability at least $1-\delta_{k-1}$, 
\begin{equation*}
    \left| \frac{1}{\Lambda} - \frac{1}{\hat{\Lambda}_k}\right| \leq \varepsilon_k = \frac{4}{\Lambda_{\min}}\sqrt{\frac{2\log(\delta_{k-1}^{-1})}{\nu_{k-1}}}
\end{equation*}
so that $\Lambda (1-\hat{\Lambda}_k\varepsilon_k) \leq \hat{\Lambda}_k $.
If $\hat{\Lambda}_k\varepsilon_k < 1$, then $\Lambda \leq \frac{\hat{\Lambda}_k}{1-\hat{\Lambda}_k\varepsilon_k}$, and we obtain another possible value for $\tilde{\Lambda}_k$ holding with probability at least $1-\delta_{k-1}$.

Then, by choosing $\tilde{\Lambda}_k = \min\left(\Lambda_{\max}, \frac{\hat{\Lambda}_k}{1-\hat{\Lambda}_k \varepsilon_k}, \hat{\Lambda}_k+\Lambda_{\max}^2 \varepsilon_k\right)$ when $\hat{\Lambda}_k\varepsilon_k < 1$ or $\tilde{\Lambda}_k = \min\left(\Lambda_{\max}, \hat{\Lambda}_k+\Lambda_{\max}^2 \varepsilon_k\right)$ otherwise, $\tilde{\Lambda}_k$ is always smaller or equal to $\hat{\Lambda}_k+\Lambda_{\max}^2 \varepsilon_k$ and $\tilde{\Lambda}_k-\Lambda\leq 2\varepsilon_{k,\Lambda}$ still holds.

\section{Discussion on the Upper Bound}
\label{section: discussion}
We provide here an intuition on the weak dependence of the regret in $S$.
As the confidence intervals on the arrival rates do not depend on the buffer size $S$,
we can expect a limited influence of the size of the system
on the regret. 
However, the influence of $S$ remains through the bias, 
which can reach higher values when the buffer size is increased.

\subsection{Dominant Terms Depending on the Regime}

Let us start with the upper bound \eqref{eq: final upper bound for PI} we obtain with PI. 
The first term $KS\frac{\Lambda_{\max} R_{\max}}{\mu_{\max}}$ is the only one that eventually can be increased by $S$. When the number of server grows sufficiently, this term can dominate the regret. However, when the number of server is infinite, $\mu_{\max}=S\mu$, and its dependence on $S$ vanishes.

The term proportional to the duration of the first episode $t_1$ rapidly becomes negligible when the number of episodes grows. 
% The choice of $t_1$ is arbitrary, but far from trivial.
% If we decrease $t_1$ while keeping $T_K$ constant by increasing $K$, this term decreases, but all the terms proportional to $K$ increase, meaning that an optimal balance could be found.
% However, in practice, decreasing $t_1$ increases the probability of having $\tilde{\Lambda}_2 > \lambda_{\max}$ for all job classes $i$. In this case, the policy will not be updated as the confidence intervals for the arrival rates do not change.
% Ideally, we may want $t_1$ to be chosen such that when the first episode finishes, at least one job class $i$ sees an update in the confidence interval of its arrival rate. 
% Waiting longer means missing an immediate improvement, but also gathering more data for finer confidence intervals.
% Furthermore, if no confidence interval is updated at $t_1$, we need to wait just as long for the next possible update. 

% The two terms proportional to $Km\log T_K$ are dominated by the term in $O(m\sqrt{T_K \log T_K})$ and therefore will never be predominant.

The term in $O(\left(\Lambda_{\max}^2/\Lambda_{\min}+\sqrt{m}\right)\sqrt{T_K \log T_K})$ will be the dominant one in two regimes: when $m$ grows or when the total time $T_k$ increases sufficiently.
It also highlights that we want $\Lambda_{\min}$ and $\Lambda_{\max}$ to be as tight as possible.

The remaining terms are negligible if $S$ or $T_K$ grows.

For VI, the main difference comes from the introduction of $V$ given by \eqref{eq: expression of V}. In the worst case, it adds an exponential dependence on the queue size $S$, which multiplies in particular the term proportional to $\sqrt{T_K\log T_K})$. 
Yet, this term results more from a lack of guarantees than from the certainty of a bad behavior. 
Empirically, VI does not seem to differ much from PI as for all our experiments both methods yield the exact same regret. 
In terms of convergence speed, in our experiments PI required on average $4.2$ iterations and $9.3$ms (within $0.1$ ms) of running time to converge for the episodes updating the optimistic arrival rates. VI required $1217$ iterations on average and $1.3$s (within $0.1$ s) of running time.

% Let us finish this section with a remark on the influence of the bounds $\lambda_{\min}$ and $\lambda_{\max}$ on the arrival rates of the optimistic CTMDP.
% The optimistic rate for class $i$ and episode $k\geq 2$ is $\tilde{\lambda}_k^{(i)}=\min\left(\lambda_{\max}, \hat{\lambda}_k^{(i)}+\varepsilon_k^{(i)}\right)$, with $\varepsilon_k^{(i)}$ defined in proposition \eqref{proposition: confidence intervals}. Since $\varepsilon_k^{(i)}$ is proportional to $\lambda_{\max}^2 / \lambda_{\min}$, increasing $\lambda_{\max}$ or decreasing $\lambda_{\min}$ delays the time at which the optimistic arrival rates becomes smaller than $\lambda_{\max}$. 
% Thus, the less precisely $\lambda_{\min}$ and $\lambda_{\max}$ bounding the arrival rates are known, the longer the waiting time before optimistic arrival rates become smaller than $\lambda_{\max}$. Similarly, more dispersed the arrival rates lead to a larger gap between $\lambda_{\min}$ and $\lambda_{\max}$ and result in the same slower convergence. 

\subsection{Comparison with other Algorithms}
The best regret upper bound for general algorithms on undiscounted infinite horizon MDPs is of order $\tilde{O}\left(\sqrt{DXAT}\right)$, where we recall that $D$ is the diameter of the MDP, $X$ the size of the state space, $A$ the size of the action space and $T$ the time.
By exploiting the structure of the problem, 
our method achieves a bound that does not depend on the MDP diameter $D$ and asymptotically does not depend on the size of the state space $X=S+1$,
though boundaries on the global arrival rate are required.
% The main difference between Jaksch et al's result and ours is that their regret bound does not require boundaries on the arrival rates, whereas ours does not depend on the diameter.

As we know the values $r^{(i)}(s)$, we can establish for any state the priority of each job class, so that we can reduce the action space to $m+1$ actions by defining action $a$ as accepting the $a$ highest-priority job classes.
Thus, the number of classes $m$ appears under a square root in the state-of-the-art bounds, as it does in our bound.

With PI, for the regime in which the time $T$ dominates and if the immediate rewards $R^{(i)}, i=1, \ldots, m$ are bounded, we obtain an upper bound with a main term of $O(\sqrt{mT\log T})$ that does not depend on $S$,
similarly to Anselmi et al. \citeyearpar{anselmiReinforcementLearningBirth2022} and their bound $\tilde{O}(\sqrt{E_2 A T})$, where $E_2$ is a coefficient independent on $S$. 
A lack of guarantees on the relative bias prevents this result to hold for VI.
Indeed, the policy obtained by VI satisfies \eqref{eq: average reward for VI} and its optimality is not certain.
Thus, some results of Feinberg and Yang bounding the relative bias $\nabla h$ for gain optimal policy cannot be applied.
For more details, see the proof of lemma \ref{lemma: Rgood} in appendix \ref{appendix: regret}.

\section{Time and Space Complexity}
\label{annex: time and space complexity}
\paragraph{PI}
PI requires a memory space of size $S$ to store the values of the average reward $\rho^\pi$ and the relative bias $\nabla h^\pi$. The policy can be represented by an array of size $S$ using the admission priorities as discussed in section \ref{section: regret}. The computation of the average reward $\rho^\pi$ and the relative bias $\nabla h^\pi$ with the naive method using the LAPACK DGESV routine requires the use of the infinitesimal generator $Z^\pi$, i.e. an additional memory space of $O(S^2)$. 
The method that we propose in Section \ref{section: computing the bias} requires only a memory space of $O(S)$.
% Therefore, PI uses a total memory space of size $O(S^2)$. 

Each policy evaluation costs $O(S^3)$ computations as explained in section \ref{section: computing the bias} with the LAPACK DGESV routine, or $O(S)$ according to Section \ref{section: computing the bias}, and must be performed at most $(m+1)^S$ times in the worst case. To the best of our knowledge, no bounds on the number of iterations have been proposed for the average reward setting. For the improvement step, $O(mS)$ comparisons are run. The total time complexity per iteration is thus $O(S^3 + mS)$ for the naive approach or $O(mS)$ with our proposition in Section \ref{section: computing the bias}.

\paragraph{VI}
The bias vector $u$ tries to approximate the relative bias $\nabla h^\pi$. The transition probability matrix $P_U^\pi$ of the uniformized problem is used instead of the infinitesimal generator $Z^\pi$. The total memory usage is $O(S^2)$.

VI converges geometricaly fast and each iteration requires $O(S^2)$ computations for the evaluation and $O(mS)$ for the improvement. Each loop requires therefore $O(S^2 + mS)$ computations.

\paragraph{UCRL-AC}
Step 1 requires a memory space of size $O(1)$ to store $\varepsilon_{k,\Lambda}$ and $\varepsilon_{k,p}$. 
It costs $O(1)$ computations.

Step 2 has for the policy search the same time and memory complexities as the algorithm updating the policy, i.e. PI or VI.
To determine the optimistic CTMDP, we need to find the best plausible state-dependent arrival rates, which costs $O(mS)$ in time and space complexity.

For step 3, a memory space $O(m)$ is needed to estimate the arrival rates. $O(1)$ operations are performed at each arrival.

Consequently, the time and space complexity of UCRL-AC is of the same order of magnitude as the policy search algorithm that is employed. UCRL-AC is run for a total duration $T_K=t_1 2^{K-1}$ for $K$ episodes and initial episode length $t_1$, during which there are at most $\sum_i \lambda^{(i)} T_K$ arrivals and $\mu(S)T_K=c\mu T_k$ departures, i.e. at most $O(( \Lambda + c\mu )T_K)$ events occurring.

\section{Computing Infrastructure}
\label{annex: computing infrastructure}
The experiments were run on a MacBook Pro 2021 equipped with an Apple M1 Pro processor and 16 GB RAM.

%%%%%%%%%%%%%%%%%%%%%%%%%%%%%%%%%%%%%%%%%%%%%%%%%%%%%%%%%%%%%%%%%%%%%%%%%%%%%%%
%%%%%%%%%%%%%%%%%%%%%%%%%%%%%%%%%%%%%%%%%%%%%%%%%%%%%%%%%%%%%%%%%%%%%%%%%%%%%%%

\end{document}